\documentclass[journal]{IEEEtran}
\usepackage{amsmath,amsfonts}
\usepackage{algorithm}
\usepackage{algorithmic}
\usepackage{array}
\usepackage[caption=false,font=normalsize,labelfont=sf,textfont=sf]{subfig}
\usepackage{textcomp}
\usepackage{stfloats}
\usepackage{url}
\usepackage{verbatim}
\usepackage{graphicx}

\usepackage{cite}
\usepackage{algorithm}
\usepackage{amssymb}
\usepackage{multirow}
\usepackage{booktabs}
\usepackage{siunitx}
\usepackage{pifont}
\usepackage{orcidlink}
\usepackage{threeparttable}
\usepackage{makecell}
\usepackage{colortbl}
\usepackage{xcolor}

\def\BibTeX{{\rm B\kern-.05em{\sc i\kern-.025em b}\kern-.08em
    T\kern-.1667em\lower.7ex\hbox{E}\kern-.125emX}}
\usepackage{balance}

\begin{document}
\title{Leveraging Event Cameras for Robust Exposure Correction: \\A New Real-world Dataset and Method}
\author{Siru Zhang\,\orcidlink{0009-0006-2764-5863}, 
Huiyuan Fu\textsuperscript{*}\,\orcidlink{0000-0002-5276-4366},~\IEEEmembership{Member,~IEEE,} 
Qinghua Yang\,\orcidlink{0009-0000-3887-8757}, 
Hao Kang\,\orcidlink{0009-0007-8156-9045}, 
Zekai Xu\,\orcidlink{0009-0003-3386-0701}, \\
Yaoyao Zhong\,\orcidlink{0000-0002-2671-9350},~\IEEEmembership{Member,~IEEE,} 
Xin Wang\,\orcidlink{0000-0001-8639-3818},~\IEEEmembership{Senior Member,~IEEE,} 
Huadong Ma\,\orcidlink{0000-0002-7199-5047},~\IEEEmembership{Fellow,~IEEE,}
\thanks{
This work is supported in part by the NSFC under No.62272059 and No.U24B20176, the National Key R\&D Program of China under No.2023YFF0904800, the Beijing Natural Science Foundation under No.JQ24020, and the Beijing Nova Program under No.20230484406.

Siru Zhang, Huiyuan Fu, Qinghua Yang, Hao Kang, Zekai Xu, and Huadong Ma are with the State Key Laboratory of Networking and Switching Technology, Beijing University of Posts and Telecommunications, Beijing 100876, China.

Yaoyao Zhong is with Beijing Normal University, Beijing 100875, China. 

Xin Wang is with the Department of Electrical and Computer Engineering, Stony Brook University, NY 11794, USA.

*Corresponding Author: Huiyuan Fu (email: fhy@bupt.edu.cn)
}}

\markboth{IEEE Transactions on Multimedia, Vol.~1, 2026}
{Siru Zhang, 
\MakeLowercase{\textit{(et al.)}}: Leveraging Event Cameras for Robust Exposure Correction: A New Real-world Dataset and Method}
\maketitle

\begin{abstract}
Existing exposure correction datasets, typically captured using conventional low dynamic range sensors, often suffer from irreversible information loss in severely under- or over-exposed regions. This limitation hinders the accuracy and robustness of image reconstruction, particularly in complex real-world scenarios. To address this challenge, we leverage the high dynamic range sensing capability of event cameras to better capture illumination variations under extreme exposure conditions. We introduce REEC, a large-scale, real-world dataset containing 11,100 spatially and temporally aligned event-image pairs, collected using a precisely calibrated, high-resolution hybrid-camera system. The dataset covers a wide range of exposure levels, scene types, and lighting conditions, serving as a resource for event-driven low-level vision research and applications. Building upon this dataset, we propose EvENet, an event-guided exposure correction network that employs exposure-aware cross-modal masks and a similarity-driven fusion strategy with information-theoretic motivation to integrate image and event modalities. Extensive experiments across diverse datasets demonstrate that our method significantly improves detail restoration under challenging lighting conditions, advancing the state of the art in event-guided exposure correction. Our project is available at \url{https://github.com/zhangisland/EvENet}.
\end{abstract}


\section{Introduction}
\label{sec:intro}
\begin{figure}[!tp]
    \centering
    \includegraphics[width=1.0\linewidth]{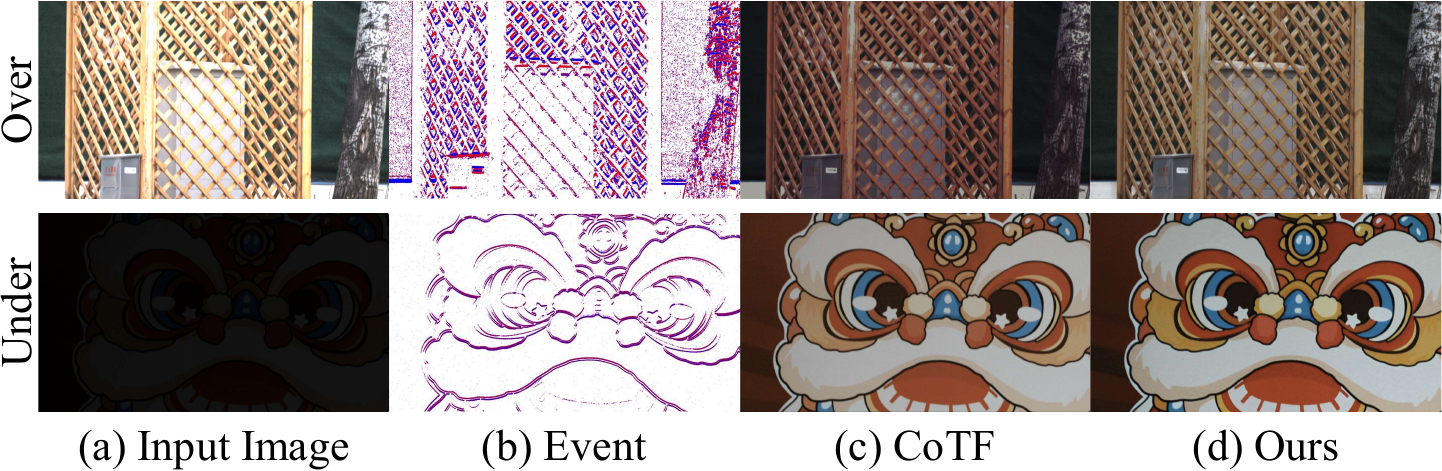}
    \caption{Given input ill-exposed images (a) along with event data (b), our method (d) leverages event as auxiliary input to preserve finer details in both regions, compared with the image-based method CoTF~\cite{liRealTimeExposure2024} (c).}
    \label{fig:intro}
\end{figure}

\IEEEPARstart{E}{xposure} correction is a crucial task in low-level vision, focusing on recovering visually informative content from under- or over-exposed images. 
Unlike low-light enhancement, which primarily targets under-exposed inputs, exposure correction requires bidirectional exposure adjustment to restore both under- and over-exposed regions, potentially within the same image. These regions require different correction behaviors, making region-specific processing important for preserving visual fidelity.
Most existing real-world datasets for exposure correction~\cite{caiLearningDeep2018,wangLocalColor2022} are primarily collected using low dynamic range (LDR) sensors.
Images captured by these sensors inherently struggle to preserve structural and intensity details in both under- and over-exposed regions simultaneously.
Recent deep learning-based methods~\cite{Cui_2022_BMVC,baekLuminanceawareColor2023b,zhou4KResolutionPhoto2024,liuRegionAwareExposure2024,liuLearningExposure2024,liRealTimeExposure2024,liColorShift2024,jungExposureslotExposurecentric2025a} have shown significant advances in non-uniform illumination image exposure correction by learning mappings from poorly exposed inputs to well-exposed outputs. 
However, limited by existing datasets, these methods still fail to perform well under challenging lighting conditions, as shown in Fig.~\ref{fig:intro}.
Another related orientation of addressing above issues is Multi-exposure image fusion (MEF)~\cite{wuACEMEFAdaptive2023,wuHybridSupervisedDualSearch2024,zhuTaskCustomizedMixture2024} and HDR imaging~\cite{leSingleImageHDR2023,yangLearningEvent2023,chenUltraFusionUltra2025b,zouEventHDREvent2025a,duanEventAidBenchmarking2025}. However, such methods primarily rely on multi-exposure inputs, which limit the application of these methods in real-world scenarios.

Event cameras~\cite{gallegoEventBasedVision2022} offer high dynamic range, microsecond-level temporal resolution, and low latency, making them particularly well-suited for low-level vision tasks.
These sensors asynchronously capture per-pixel illumination changes, overcoming the inherent limitations of conventional sensors by providing wider dynamic ranges and precise temporal information.
However, events encode brightness changes rather than absolute intensity or color, and their usefulness for exposure correction varies across image regions. Effective event-guided exposure correction requires selective integration of event cues to support bidirectional exposure adjustment.

These considerations motivate us to focus on two key challenges: (i) the need for paired real-world RGB-event data covering both under- and over-exposure for systematic evaluation, and (ii) the design of a cross-modal integration framework that selectively incorporates event cues for region-specific correction and jointly considers cross-modal alignment and multi-scale feature diversity.
Our main contributions are twofold:

\noindent\textit{(i) Dataset.} 
We introduce REEC, a Real-world Event-based Exposure Correction dataset comprising 11,100 spatially and temporally aligned event-image pairs captured under both under- and over-exposed conditions.
This dataset is collected using a high-resolution hybrid-camera system with precise calibration and alignment, covering a wide range of scenes and lighting scenarios. 
Compared to existing real-world datasets (as shown in Tab.~\ref{tab:dataset_comparison}), REEC provides richer modalities and a broader range of exposure distributions.

\noindent\textit{(ii) Method.} 
Built upon REEC, we propose EvENet, an event-guided exposure correction network designed for robust performance in real-world scenarios.
The Cross-modal Mask Exposure Aware (CMEA) module uses an image-derived SNR mask and its complement to guide separate branches for under- and over-exposure, while an event-derived activity mask guides multi-scale features through cross-mask enhanced aggregation.
The Cross-Modal Alignment and Fusion (CMAF) module introduces a joint similarity-based learning objective that encourages image-event feature alignment at corresponding scales and discourages redundant image representations across adjacent scales.
These modules combine region-specific exposure correction with selective event guidance and joint regularization of cross-modal and inter-scale features.
Extensive experiments on REEC and other datasets demonstrate that EvENet significantly outperforms existing state-of-the-art methods, particularly under challenging lighting conditions.

Our main contributions are as follows:
\begin{itemize}
    \item  We design a hybrid-camera system to construct REEC, a comprehensive real-world event-based dataset aimed at exposure correction. It contains 11,100 rigorously aligned event-image pairs. The dataset features hardware-synchronized acquisition for precise temporal alignment and employs a cross-modal spatial alignment strategy.
    \item We propose a novel Event-guided Exposure Correction Network (EvENet) that adaptively fuses event and image modalities via cross-modal masking and a dual-objective feature regularization strategy, which enforces cross-modal alignment and inter-scale decorrelation to preserve high-fidelity details under real-world scenarios.
    \item We conduct extensive experiments to demonstrate the effectiveness of the REEC dataset and the performance of EvENet. Results show that our method consistently outperforms existing state-of-the-art approaches.
\end{itemize}

\section{Related Works}
\label{sec:related_works}
\subsection{Datasets}
\subsubsection{Image-based Datasets}
Most existing exposure correction datasets are built exclusively using image-based modalities from conventional cameras.
Representative datasets include SICE~\cite{caiLearningDeep2018} and LCDP~\cite{wangLocalColor2022}, which provide paired poorly exposed images with corresponding refined ground truth references. While these datasets have enabled progress in supervised learning, their passive imaging nature limits the recovery of structural details in severely under- or over-exposed regions, highlighting the need for complementary sensing modalities.

\begin{table}[!t]
    \begin{threeparttable}
    \caption{Comparison of REEC with existing real-world datasets. REEC bridges the gap in event-based exposure correction by using synchronized RGB and event data.}
    \label{tab:dataset_comparison}
    \centering
    \begin{tabular}{lccccr}
    \toprule
    Dataset & Modality & Exposure & {\#Pairs} & Resolution\\
    \midrule
    SICE (2018) & I & U+O &4,413 & Varies\\
    LCDP (2022) & I & U+O &1,733 & 1920$\times$1080 \\
    LIE (2024) & I+E & U  &2,231 & 256$\times$256 \\
    SDE (2024) & I+E & U  &31,477 & 346$\times$260\\
    \midrule
    REEC (Ours) & I+E & U+O &11,100 & 1280$\times$720 \\
    \bottomrule
    \end{tabular}
    \begin{tablenotes}
        \item ``I'' and ``E'' denote image and event modalities. ``U'' and ``O'' indicate datasets containing underexposed and overexposed pairs.
    \end{tablenotes}
    \end{threeparttable}
\end{table}

\subsubsection{Event-based Datasets}
Although event cameras offer high dynamic range (HDR) capabilities, existing event-based datasets are not specifically tailored for exposure correction.
Current datasets primarily fall into two categories: 

\textit{(i) HDR reconstruction datasets}, such as EventAid~\cite{duanEventAidBenchmarking2025} and EventHDR~\cite{zouEventHDREvent2025a}, generate HDR ground truths by merging multiple LDR images. 
However, when these datasets are tone-mapped to the standard dynamic range (SDR) for exposure correction or evaluation, artifacts and bias may be introduced.

\textit{(ii) Low-light enhancement datasets}~\cite{liangCoherentEvent2023,jiangEventBasedLowIllumination2024,liuSeeingMotion2024,liangRobustEventguided2024} focus solely on low-light scenarios, thus lacking the illumination diversity necessary for general exposure correction.

These limitations highlight the need for our REEC dataset, explicitly designed for event-based exposure correction, encompassing both under- and over-exposed scenarios.

\subsection{Image-Only Methods}
Image enhancement is predominantly addressed by image-only methods as the image modality is easy to obtain. From the target perspective, existing methods can be divided into two aspects: Low-light enhancement and exposure correction.

\subsubsection{Low-light Enhancement Methods}
Image-based low-light enhancement methods~\cite{liLearningEnhance2022,changLowLightImage2022, fuLearningSimple2023,caiRetinexformerOnestage2023,maLowLightImage2023,zhengUnsupervisedUnderexposed2023} aim to improve the visibility and perceptual quality of underexposed images through various learning paradigms.
Representative approaches include curve-based optimization~\cite{liLearningEnhance2022}, Retinex-based modeling~\cite{fuLearningSimple2023,caiRetinexformerOnestage2023}, and learning illumination priors or self-supervised guidance~\cite{maLowLightImage2023,zhengUnsupervisedUnderexposed2023}.
While effective in low-light scenes, these methods struggle to generalize to diverse exposure conditions.
In contrast, our method achieves exposure correction under diverse illumination conditions by integrating image-dominant learning with event-guided supervision.

\subsubsection{Exposure Correction Methods}
Existing exposure correction methods predominantly utilize image-based approaches due to the lack of multimodal datasets. 
These methods can be broadly categorized into two groups:

\textit{(i) Holistic methods} process images as a whole. Traditional methods~\cite{wangExperimentBasedReview2020b} include histogram equalization and gamma correction, while learning-based methods estimate ISP parameters~\cite{Cui_2022_BMVC} or decompose images into multi-scale representations~\cite{zhou4KResolutionPhoto2024}. 
Although effective for uniformly exposed images, these methods struggle with non-uniform exposure variations.

\textit{(ii) Region-aware methods}~\cite{wangLocalColor2022,liColorShift2024,liRealTimeExposure2024,liuRegionAwareExposure2024,jungExposureslotExposurecentric2025a} tackle non-uniform exposure challenges by learning region-specific representations, thereby enabling spatially adaptive enhancement.
For example, LCDP~\cite{wangLocalColor2022}, LACT~\cite{baekLuminanceawareColor2023b}, and CSEC~\cite{liColorShift2024} exploit relative color shifts and apply color transformation strategies to correct regional exposure discrepancies.

Beyond single-image methods, multi-exposure fusion (MEF) methods~\cite{wuACEMEFAdaptive2023,wuHybridSupervisedDualSearch2024,zhuTaskCustomizedMixture2024} combine both under- and over-exposed images to improve exposure correction, while HDR approaches~\cite{leSingleImageHDR2023,kongSAFNetSelective2024,liAFUNetCrossIterative2025} fuse multiple LDR exposures to produce HDR images.
However, these methods generally rely on well-aligned, high-quality data and require tone mapping for standard display, which limits their practical applicability.

\subsection{Event-Only Methods}
With the development of the event camera, the superiority of high dynamic range and high temporal resolution makes it more suitable for video reconstruction, deblurring, and interpolation tasks than traditional LDR color cameras.
Existing event-only reconstruction methods primarily rely on intensity changes to reconstruct images or videos. The event stream encodes continuous illumination variations, which can also be used to infer grayscale texture information.

However, event streams generally contains noise. Recent advances address event signal quality and representation learning, which include probability-flux-based filtering for improved signal fidelity~\cite{chenEventStream2026} and functional backpropagation for unbiased gradient estimation through event binning~\cite{chenUnbiasedGradient2026}.
These advances improve event signal fidelity and representation learning, supporting the effective use of event information in visual tasks. Our work explores event-image complementarity for exposure correction.
Related works tend to directly integrate event streams and map event signals to grayscale images or videos\cite{rebecqEventsToVideoBringing2019a,liuSeeingMotion2024,wuEventbasedAsynchronous2024a,liuNERNetSeeing2025}.
One milestone work, E2VID~\cite{rebecqEventsToVideoBringing2019a}, proposes an end-to-end framework to reconstruct grayscale videos from event stream. And there are some following works~\cite{wengEventBasedVideo2021,liuSeeingMotion2024,liuNERNetSeeing2025} that reconstruct the videos for low-light enhancement.
There are also other related works towards other tasks, e.g., image super-resolution~\cite{mohammadmostafaviiLearningSuper2020}, HDR imaging~\cite{wuEventbasedAsynchronous2024a}, deblurring, and interpolation.
E2SRI~\cite{mohammadmostafaviiLearningSuper2020} extends events to the image super-resolution task.
AsynHDR~\cite{wuEventbasedAsynchronous2024a} assembles an event sensor with LCD panels to modulate the irradiance incident and trigger the pixel-independent event streams, thereby achieving higher dynamic range.
However, event-only methods are inherently limited to reconstructing grayscale images, as the event stream alone does not provide sufficient information to recover color, thereby degrading the visual quality.

\noindent\textbf{Spike-based imaging.}
Spike cameras provide another neuromorphic imaging paradigm. Huang et al.~\cite{huang1000Faster2023} introduce a camera that generates spikes when accumulated incident light reaches a threshold, enabling high-speed intensity reconstruction. Building on spike sensing, USP-Gaussian~\cite{chenUSPGaussianUnifying2025} jointly addresses spike-based image reconstruction, pose correction, and Gaussian splatting. 
These studies demonstrate the potential of spike-based sensing for high-speed imaging and scene reconstruction. While spike cameras encode accumulated light through spike timing, event cameras encode temporal changes in logarithmic intensity. Our work combines events with RGB images to exploit their complementary information for exposure correction.

\subsection{Hybrid Image-Event Methods}
Image-only methods struggle to handle drastic illumination changes, while event-only methods are inherently limited to grayscale intensity changes and cannot recover color information. These complementary limitations naturally motivate the development of image-event fusion approaches, which leverage the strengths of both modalities to achieve robust visual sensing across diverse lighting conditions. 
However, despite growing interest in fusion-based methods, event-guided exposure correction remains an underexplored domain, presenting opportunities for investigation.

Existing research has primarily focused on low-light enhancement~\cite{jiangEventBasedLowIllumination2024,liangRobustEventguided2024}, HDR reconstruction~\cite{wangEventBasedHigh2019,rebecqHighSpeed2021,yangLearningEvent2023,zouEventHDREvent2025a}, and other spatiotemporal restoration tasks, e.g., deblurring~\cite{panHighFrame2022,kimEventguidedDeblurring2022,sunEventBasedFusion2022,xiaopengHDRImaging2024} and interpolation~\cite{sunUnifiedFramework2025}.
Beyond image restoration, event-based 3DGS~\cite{hanEvent3DGSEventbased2024,liao2025ef3dgs} combines events and frames to jointly optimize camera poses and 3D Gaussian scene representations. This demonstrates the value of event-frame fusion for scene reconstruction.

\subsubsection{Low-light Enhancement}
Representative event-guided low-light enhancement approaches include LIE~\cite{jiangEventBasedLowIllumination2024}, which employs residual fusion to restore fine-grained details in under-exposed regions, and EvLight~\cite{liangRobustEventguided2024}, which introduces SNR-guided feature selection for adaptive low-light enhancement.

\subsubsection{HDR Reconstruction}
Event-guided HDR reconstruction method, HDRev~\cite{yangLearningEvent2023}, proposes a multimodal representation alignment strategy to fuse image and event modalities.
However, HDR imaging methods trained on SDR datasets often perform poorly due to the dynamic range gap.
Although such models can be retrained for SDR data, their dynamic range modeling becomes ineffective when both input and supervision lack HDR information.
In contrast, our approach directly targets SDR reconstruction and leverages event signals to handle diverse exposure conditions.

\subsubsection{Other Spatiotemporal Restoration Methods}
Event streams encode asynchronous brightness changes with high temporal resolution, providing complementary spatiotemporal cues to conventional frame-based restoration.
Early event-guided methods formulate intensity recovery as an integration problem over event streams, e.g., mEDI~\cite{panHighFrame2022} regards the blurred image as the integral of a sequence of latent images, and associates the event data to a latent sharp image, thus removing deblur and reconstructing HFR videos.
UEVD~\cite{kimEventguidedDeblurring2022} considers the exposure and readout time to estimate the correlation between blurred frames and events.
EFNet~\cite{sunEventBasedFusion2022} introduces a symmetric cumulative event representation and applies multiple cross-modal attention modules for enhancing both spatial detail recovery and temporal consistency.
Self-EHDRI~\cite{xiaopengHDRImaging2024} proposes a unified self-supervised HDR and deblurring framework that learns cross-domain conversions from blurry LDR images to sharp LDR images, and leverages events to recover the intermediate sharp HDR results.
More recently, REFID~\cite{sunUnifiedFramework2025} constructs a bidirectional recurrent network for interpolation that performs deblurring ad hoc.

Despite the promising performance of existing methods on their respective tasks, such as deblurring, HDR reconstruction, and interpolation, their optimization objectives are inherently tailored to task-specific degradation models. 
Such formulations do not explicitly account for the illumination inconsistency and exposure bias that dominate exposure correction scenarios. Consequently, directly applying these methods may lead to suboptimal fusion behaviors. This motivates the need for a dedicated image-event fusion strategy that is explicitly aligned with the objective of exposure correction.

\subsection{Multimodal Fusion Methods}
Multimodal fusion aims to align or complement heterogeneous features across modalities.
The most intuitive approach is to leverage an attention mechanism.
EFNet~\cite{sunEventBasedFusion2022} performs motion deblurring by applying channel-level attention to fuse frame and event features, while EGEIF~\cite{zhaoEdgeGuidedFusion2024} leverages edge-guided filtering to compensate for information loss in sparse event streams.
Another line of research explores information-theoretic principles for multimodal representation learning~\cite{hjelmLearningDeep2019,zhouMutualInformationDriven2022,shenMutualInformationdriven2023b}, often formulated through contrastive objectives that maximize statistical dependency while reducing redundancy.
However, these methods typically rely on explicit density-ratio estimation or contrastive learning frameworks, and are mainly designed for unimodal representation learning or specific fusion tasks such as PAN-sharpening~\cite{zhouMutualInformationDriven2022}.

Instead, we adopt a lightweight similarity-driven formulation that is conceptually motivated by these principles, while being tailored to event-frame fusion. Specifically, we jointly enforce cross-modal alignment and inter-scale decorrelation, which proves to be both effective and computationally efficient for exposure correction under challenging lighting conditions.

\begin{figure*}[!tp]
    \centering
    \includegraphics[width=1.0\linewidth]{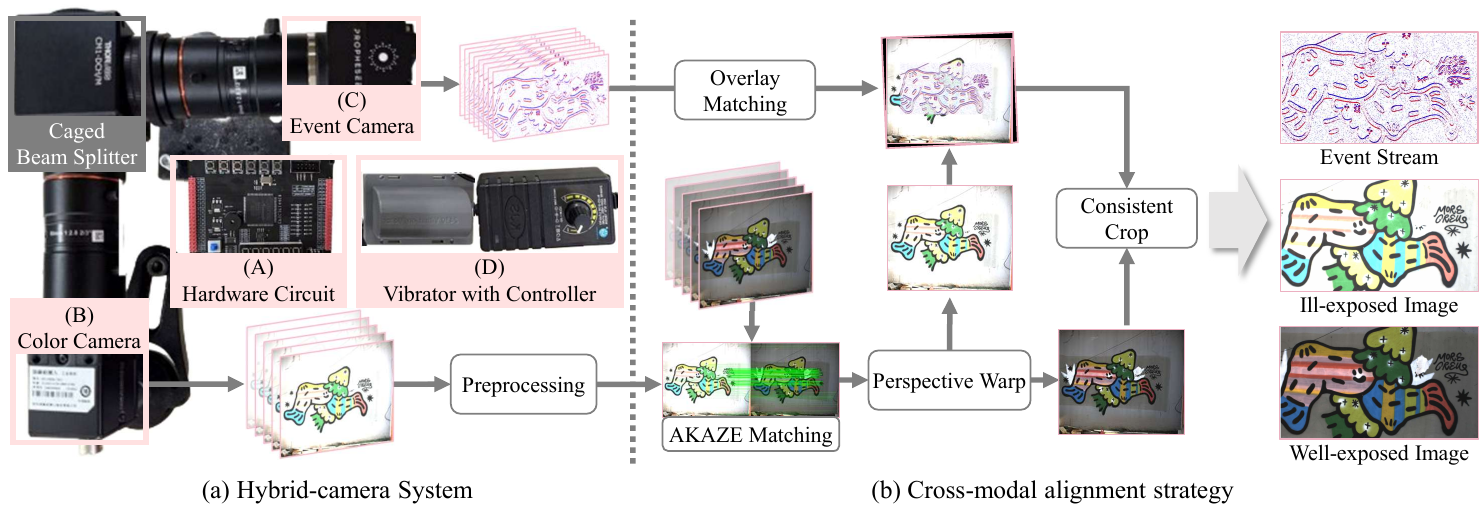}
    \caption{The dataset collection pipeline. (a) Our hybrid-camera imaging system collects diverse exposure pairs. (b) Data processing ensures highly aligned event-image pairs.}
    \label{fig:camera_system}
\end{figure*}

\section{Our Proposed Dataset}
\label{sec:dataset}
Existing exposure correction datasets suffer from abrupt exposure variations that fail to reflect real-world continuous lighting transitions, thereby hindering effective event streams synthesis.
To address this limitation, we introduce REEC, a large-scale Real-world Event-based Exposure Correction dataset, constructed using a precisely calibrated hybrid-camera system.
REEC contains 11,100 spatially and temporally aligned event-image pairs captured under diverse and realistic exposure conditions, enabling robust event-based exposure correction and advancing further research.

\begin{figure*}[!t]
    \centering
    \includegraphics[width=1.0\linewidth]{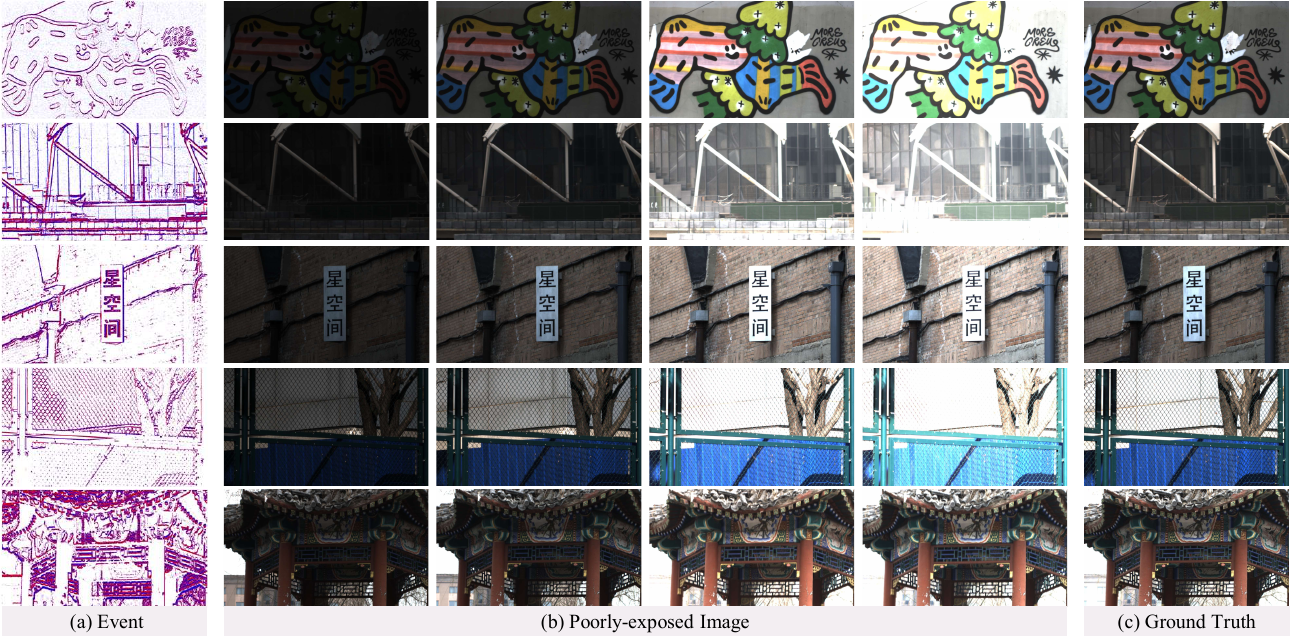}
    \caption{Data samples from our proposed REEC dataset.}
    \label{fig:vis_sample}
\end{figure*}

\subsection{Apparatus}
To collect the REEC dataset, we design a sophisticated hybrid-camera system, as illustrated in Fig.~\ref{fig:camera_system}(a). 
The system consists of two cameras equipped with a beam splitter: one captures 2448$\times$2048 color frames, while the other records 1280$\times$720 event streams.
Unlike the classical DAVIS346, which produces low-resolution and noisy color outputs, our system can ensure the acquisition of a high-resolution dataset.

To ensure precise temporal alignment between event streams and color frames, we implemented a hardware-based synchronization mechanism (A) that sends trigger signals to both the color camera (B) and the event camera (C) to start recording simultaneously, obviating the need for manual temporal alignment.
Since event cameras only generate events in response to motion or illumination changes, we use a motorized vibration mechanism (D) to induce subtle, controlled motion in the event camera. 
This approach produces consistent, temporally aligned event streams while also introducing slight motion that simulates realistic camera shake.

\begin{figure}[!tbp]
    \centering
    \includegraphics[width=1.0\linewidth]{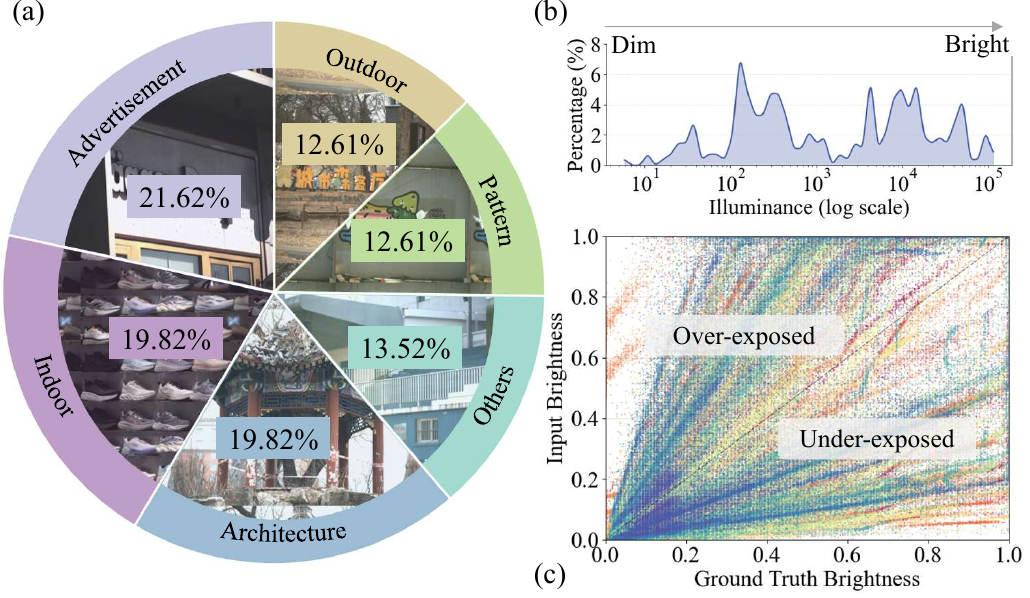}
    \caption{Statistical analysis of the REEC dataset: (a)-(b) distributions of scene categories and scenes illuminance, and (c) brightness mapping curves.}
    \label{fig:overview_dataset}
\end{figure}

\begin{figure}[!tbp]
    \centering
    \includegraphics[width=1.0\linewidth]{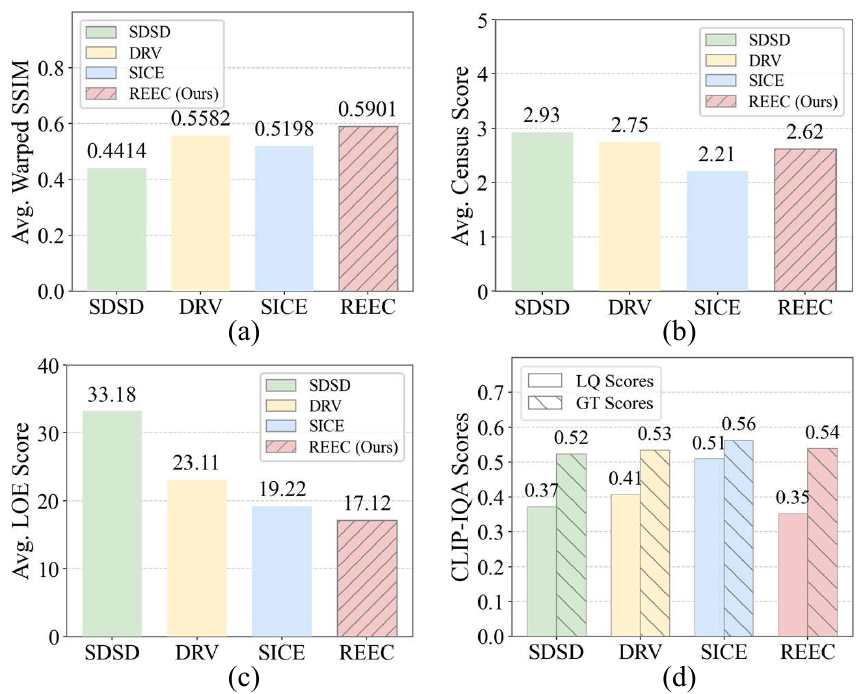}
    \caption{Dataset quality comparison. (a)-(b) Structural consistency evaluated using warped SSIM$\uparrow$ and census scores$\downarrow$. (c) Relative lightness-order preservation measured by LOE$\downarrow$. (d) Image quality assessed by CLIP-IQA$\uparrow$.}
    \label{fig:reec_alignment_analysis}
\end{figure}

\subsection{Data Collection}
For each scene, we capture four sequences under varying exposure levels, along with corresponding normally exposed images. The exposure levels are controlled by adjusting the exposure time, which ranges from 1 ms to 500 ms depending on the scene illumination.
We vary frequency and exposure settings to produce a wide range of motion magnitudes rather than a single fixed periodic pattern.
For challenging scenes where it is difficult to directly capture normally exposed images, we involve professional photographers to manually adjust the images to produce ground-truth references to ensure perceptually consistent images, which aligns with widely adopted practices in datasets such as MIT-Adobe FiveK and SICE.
Finally, we collected data from 111 scenes, each captured at four different exposure levels, yielding a total of 11,100 event-image pairs.

\subsection{Alignment Strategy}
As illustrated in Fig.~\ref{fig:camera_system}(a), several components of our hybrid-camera system require manual assembly, such as integrating the beam splitter with the lens. 
This assembly process may introduce minor inaccuracies, leading to slight spatial misalignments in the captured data. 
Additionally, due to the distinct optical formats of the sensors used in the two cameras, their respective fields of view may differ, further complicating alignment. These challenges necessitate careful postprocessing to spatially align the captured image and event pairs.

Traditional feature-based alignment methods are ineffective in this setting due to significant illumination variations and the inherent modality gap between RGB images and event representations. To address these challenges, we develop a cross-modal alignment strategy, as shown in Fig.~\ref{fig:camera_system}(b).
Our approach uses feature matching for color frames alignment and overlay-based strategies for event-frame alignment.
The final step employs cropping the color frames using transformation parameters to achieve spatial correspondence with the event data.
All resulting data are further verified through manual inspection to ensure high quality. Representative samples are shown in Fig.~\ref{fig:vis_sample}.

\subsection{Analysis of Our REEC Dataset}
The REEC dataset covers a wide range of real-world scenes under diverse illumination conditions.
We use the DELIXI illuminometer to measure the lux value of the scene. 
The lux values span from 5.3 to 127,500 lux.
Notably, scenes with illuminance below 10 lux correspond to extremely low-light conditions, where event signals exhibit inherently degraded signal-to-noise ratios.
Our dataset is constructed from two complementary aspects. First, brightness variation is introduced by controlling the exposure time to simulate under- and over-exposure conditions. Second, we collect scenes under diverse real-world illumination levels, ensuring coverage from extremely low-light to bright environments.
A comprehensive statistical analysis of REEC is conducted to evaluate its diversity and quality.  
Fig.~\ref{fig:overview_dataset}~(a) and (b) show the scene distribution and illuminance levels, indicating broad coverage of exposure conditions.
Following~\cite{wangLocalColor2022}, brightness distributions are computed by randomly sampling pixels from both the input and ground truth images.  
As shown in Fig.~\ref{fig:overview_dataset}(c), each curve represents the brightness distribution of an individual data pair in REEC.
To assess dataset quality, we evaluate structural consistency, relative lightness-order preservation, and image quality. The census-transform-based score~\cite{zabihNonparametricLocal1994a} measures local structural differences using the average Hamming distance after the census transform, while warped SSIM evaluates structural similarity after warping. Lightness Order Error (LOE)~\cite{wangNaturalnessPreserved2013} measures relative lightness-order preservation, and CLIP-IQA~\cite{wang2022exploring} assesses image quality.

As shown in Fig.~\ref{fig:reec_alignment_analysis}, REEC achieves the highest warped SSIM and the second-lowest census score, indicating competitive structural consistency. SICE achieves a lower census score, benefiting from static multi-exposure acquisition, whereas REEC involves camera motion during acquisition and requires subsequent spatial alignment. These quantitative assessments are complemented by manual inspection to verify the quality of the aligned pairs.
REEC also achieves the lowest LOE, indicating better preservation of relative lightness order. Our GT achieves higher perceptual quality scores compared to the corresponding low-quality inputs and is competitive with existing datasets.
These results complement the structural analysis by characterizing the brightness-order consistency and image quality of REEC.
The dataset is divided into 89 training scenes (8,900 pairs) and 22 testing scenes (2,200 pairs).

\begin{figure*}[!t]
    \centering
    \includegraphics[width=\linewidth]{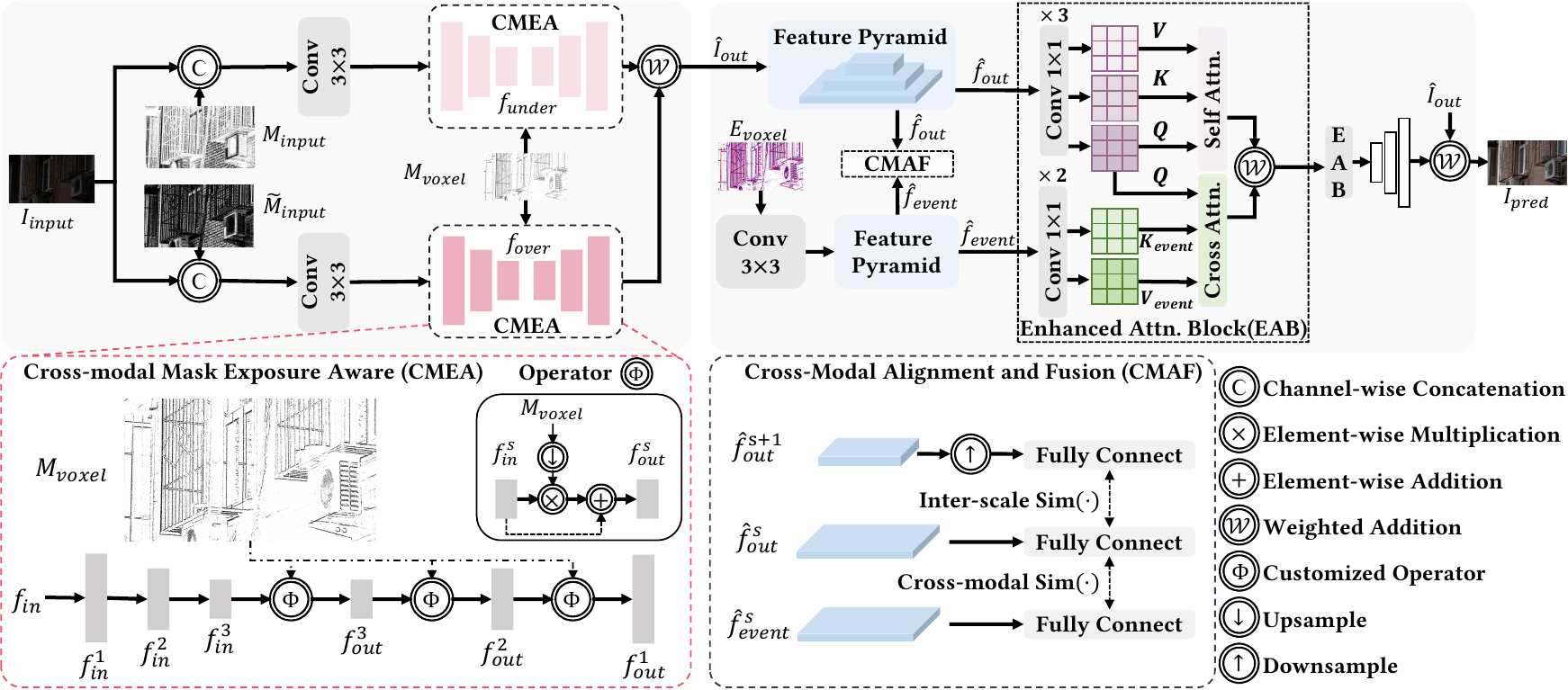}
    \caption{Overview of the proposed Event-guided Exposure Correction Network (EvENet). Given an ill-exposed image $\mathbf{I}_{input}$, we generate an SNR mask $\mathbf{M}_{input}$ and its reverse $\mathbf{\tilde{M}}_{input}$ to identify under- and over-exposed regions. An event-guided mask $\mathbf{M}_{voxel}$ is derived from voxel-based events $\mathbf{E}_{voxel}$. Dual CMEA modules extract exposure-aware features using these masks. The Cross-modal Alignment and Fusion (CMAF) module integrates image and event features, while stacked Enhanced Attention Blocks (EABs) refine representations to produce $\mathbf{I}_{pred}$.}
    \label{fig:model_arch}
\end{figure*}

\section{Methodology}
\label{sec:methodology}
Event-based exposure correction remains largely unexplored. Existing event-based methods primarily focus on HDR reconstruction~\cite{liGeneralizingEventbased2024,cuiColorEvent2024}, which aims to expand dynamic range by fusing multiple exposures, or low-light enhancement~\cite{liangCoherentEvent2023, caiRetinexformerOnestage2023,fuYouNot2023,jiangEventBasedLowIllumination2024}, targeting under-exposed scenes. 
In contrast, traditional image-based exposure correction methods primarily rely on Retinex theory~\cite{land1977retinex} or multiple exposure estimation~\cite{liuLearningExposure2024}. However, Retinex-based approaches require additional priors for illumination-reflectance decomposition and fail in over-exposed regions where illumination saturates, resulting in irreversible information loss.

To address these limitations, we leverage event cameras that asynchronously capture pixel-level brightness changes, providing complementary temporal information in challenging exposure conditions. Each event is represented as a tuple $(t, x, y, p)$, where $t$ denotes the timestamp, $(x, y)$ represents the pixel coordinates, and $p \in \{-1, +1\}$ indicates the polarity of intensity change. Following~\cite{rebecqEventsToVideoBringing2019a}, we represent these events as voxel grids $\mathbf{E}_{voxel}\in \mathbb{R}^{H\times W\times C}$, where $C$ denotes the number of temporal bins.

Inspired by~\cite{liuLearningExposure2024}, we implement a dual-branch exposure estimation module integrating cross-modal masks to obtain the over- and under-exposed regions.
We formulate the event-guided exposure correction problem as:
\begin{equation}
    \label{eq:basic_problem}
    \mathbf{I}_{pred} = \mathbf{\Gamma} (\mathbf{I}_{input}, \mathbf{E}_{voxel}),
\end{equation}
where $\mathbf{\Gamma}$ denotes our proposed Event-guided Exposure correction Network (EvENet), $\mathbf{I}_{input}$ is the ill-exposed input image, and $\mathbf{I}_{pred}$ is the corrected output. Fig.~\ref{fig:model_arch} illustrates the architecture of EvENet.

\subsection{Overall Architecture of EvENet}
We propose an event-guided exposure correction network, termed EvENet, consisting of three main components: (i) Cross-modal Mask Exposure Aware (CMEA) module that dynamically localizes exposure-sensitive regions across modalities, (ii) Cross-modal Alignment and Fusion (CMAF) module that optimizes cross-modal feature alignment through a dual-objective feature regularization strategy, and (iii) stacked Enhanced Attention Blocks (EABs) that progressively refine fused features for detail recovery.

Given the input poorly-exposed image $\mathbf{I}_{input}$, we first estimate the SNR map and then apply the threshold to achieve $\mathbf{M}_{input}$, along with its reverse mask $\mathbf{\tilde{M}}_{input} = 1-\mathbf{M}_{input}$ to differentiate under- and over-exposed regions.
We follow~\cite{xuSNRAwareLowLight2022a} to generate the $\mathbf{SNR}_{input}$. The formulation is given by Eq.~\ref{eq:gen_mask}.
\begin{equation}
    \label{eq:gen_mask}
    \mathbf{SNR}_{input}=\frac{\mathrm{mean\_filter}(\mathbf{I}_{input})}{\left|\mathbf{I}_{input}-\mathrm{mean\_filter}(\mathbf{I}_{input})\right|+\epsilon}, \quad \epsilon < 10^{-4}
\end{equation}
Based on the voxel representation $\mathbf{E}_{voxel}$, an event-derived motion mask $\mathbf{M}_{voxel} \in \{0,1\}^{H\times W}$ is generated by aggregating event responses from the first, middle, and last temporal bins, where pixel values are set to 1 if events are present within the specified temporal range, and 0 otherwise. 
The $\mathbf{M}_{voxel}$ identifies regions where event data can serve as auxiliary input to reconstruct images.

The overall pipeline is formulated as Eq.~\ref{eq:formula_pipeline}, 
where $\mathcal{O}_{CMEA}$ produces a preliminary corrected image $\mathbf{\hat{I}}_{out}$. The CMAF module then aligns features from $\mathbf{\hat{I}}_{out}$ and $\mathbf{E}_{voxel}$. Subsequently, stacked EABs progressively refine the fused features to capture both cross-modal and inter-scale dependencies, yielding the final prediction $\mathbf{I}_{pred}$.

\begin{equation}
    \label{eq:formula_pipeline}
    \begin{aligned}
        \mathbf{\hat{I}}_{out} =& \mathcal{O}_{CMEA} (\mathbf{I}_{input}, \mathbf{M}_{input}, \mathbf{\tilde{M}}_{input}, \mathbf{M}_{voxel}) \\
        \mathbf{I}_{pred} =& \mathcal{O}_{CMAF+EAB} (\mathbf{\hat{I}}_{out}, \mathbf{E}_{voxel})
    \end{aligned}
\end{equation}

Specifically, EABs progressively refine fused features through a combination of self-attention and cross-attention mechanisms to achieve more adaptive feature interaction.
Feature pyramids output $\hat{f}_{out}$ and $\hat{f}_{event}$ to EABs. For image modality, three $1\times 1$ convolutions generate query ($\mathcal{Q}$), key ($\mathcal{K}$), and value ($\mathcal{V}$). For event modality, two $1\times 1$ convolutions derive $\mathcal{K}_{event}$ and $\mathcal{V}_{event}$. Self-attention using $\mathcal{K}$ and $\mathcal{V}$ enhances intra-modal features, while cross-attention between $\mathcal{Q}$ and $(\mathcal{K}_{event}, \mathcal{V}_{event})$ captures inter-modal dependencies for better exposure reasoning. 
The final prediction $\mathbf{I}_{pred}$ is obtained by weighted fusion of the enhanced features and $\mathbf{\hat{I}}_{out}$, restoring well-exposed details.

\subsection{Cross-modal Mask Exposure Aware (CMEA) Module}
The CMEA module performs initial exposure correction using a dual-branch UNet architecture to process under- and over-exposed regions separately. 
It addresses mixed-exposure scenes using two complementary modality-specific masks.
The image-based SNR mask $\mathbf{M}_{input}$ identifies poorly exposed regions, while the event-based mask $\mathbf{M}_{voxel} \in \{0,1\}^{H\times W}$ aggregates temporal bins from the event voxel grid to detect dynamic changes that aid in recovering lost details.
We concatenate the input image $\mathbf{I}_{input}$ with each exposure mask individually, $\mathbf{M}_{input}$ for under-exposed regions and $\mathbf{\tilde{M}}_{input}$ for over-exposed regions, to extract the corresponding region-specific features ${f}_{under}$ and $f_{over}$, respectively.
\begin{equation}
    \label{eq:mask_concat}
    f_{under} = \text{E}([\mathbf{I}_{input}; \mathbf{M}_{input}]), f_{over} = \text{E}([\mathbf{I}_{input}; \mathbf{\tilde{M}}_{input}])
\end{equation}

To further enhance regional representations, we adopt a UNet architecture.
To incorporate motion-aware texture information, the event-derived motion mask $\mathbf{M}_{voxel}$ is integrated into multi-scale feature maps.
We introduce a cross-mask enhanced aggregation operator, denoted as $\Phi$, to adaptively refine features at each scale, as shown in Fig.~\ref{fig:model_arch}.
This mechanism selectively amplifies reliable information from either the image or the event stream, leading to more robust exposure-aware feature representations.

\subsection{Cross-Modal Alignment and Fusion (CMAF) Module}
Bridging the modality gap between images and events is challenging.
Instead of relying solely on attention mechanisms, we introduce a similarity-driven fusion strategy that is conceptually motivated by information-theoretic principles. Specifically, we enforce cross-modal alignment to encourage consistent representations between RGB and event features, while simultaneously introducing inter-scale decorrelation to reduce redundancy across feature hierarchies.

Our method adopts two complementary strategies:
\textit{(i) Cross-modal Alignment}: We maximize the similarity between image features ($f_{out}$) and event features ($f_{event}$) to encourage the model to learn shared and complementary representations. 
\textit{(ii) Inter-scale Decorrelation}: We minimize the similarity between image features at different scales to suppress redundant information and enhance scale-wise diversity.
These two objectives guide the fused representation to capture complementary cross-modal information while maintaining discriminative multi-scale features.

Compared with contrastive or information-theoretic estimation frameworks that require additional networks or negative sampling, we adopt a lightweight and efficient similarity-based formulation.
Specifically, we compute the cosine similarity between feature representations as a surrogate measure of feature dependency.
Given paired modalities features $X$ and $Y$, the similarity is defined as:
\begin{equation}
    \label{eq:mi_differentiable}
    \widehat{I}_{Sim}(X;Y)
    =\frac{1}{N} \sum_{i=1}^{N} 
    \left(
    \frac{\mathbf{x}_i}{\|\mathbf{x}_i\|_2} \right)^{\!\top}
    \left( \frac{\mathbf{y}_i}{\|\mathbf{y}_i\|_2} \right)
\end{equation}

To measure inter-scale similarity, the feature map $\hat{f}^{s+1}_{out}$ from the coarser scale is upsampled to match the spatial resolution of $\hat{f}^{s}_{out}$.
We then compute the similarity between $\hat{f}^{s}_{out}$ and the upsampled feature $\mathrm{Up}\Big(\hat{f}_{out}^{s+1}\Big)$.
Accordingly, the cross-modal and inter-scale objectives at scale $s$ are defined as Eq.~\ref{eq:mi_total}.

\begin{equation}
    \label{eq:mi_total}
    \begin{aligned}
        \hat{I}_{modal}^{(s)}&=\hat{I}\Big(\hat{f}_{out}^s;\hat{f}_{event}^s\Big) \\
        \hat{I}_{scale}^{(s)}&=\hat{I}\Big(\hat{f}_{out}^s;\mathrm{Up}\Big(\hat{f}_{out}^{s+1}\Big)\Big)\\      
    \end{aligned}
\end{equation}

\begin{table*}[ht]
    \begin{threeparttable}
    \centering    
    \caption{Quantitative results on our real-world REEC dataset. Best and second-best results are \textbf{bolded} and \underline{underlined}, respectively. Metrics on Under/Over are computed on under-/over-exposed samples. $\uparrow$/$\downarrow$ denotes that higher/lower values are better.}
    \label{tab:result_REEC}
    \begin{tabular}{ccc|S[table-format=2.2,mode=text]cc|ccc|S[table-format=2.2,mode=text]ccc}
    \toprule
    \multirow{2}{*}{Input} & \multirow{2}{*}{Method} & \multirow{2}{*}{Publication}& \multicolumn{3}{c|}{Under} & \multicolumn{3}{c|}{Over} & \multicolumn{4}{c}{Average} \\
     &&& PSNR$\uparrow$ & SSIM$\uparrow$ & LPIPS$\downarrow$ & PSNR$\uparrow$ & SSIM$\uparrow$ & LPIPS$\downarrow$ & PSNR$\uparrow$ & SSIM$\uparrow$ & LPIPS$\downarrow$ & PIQE$\downarrow$\\
    \hline
    \multirow{11}{*}{Image}&ZeroDCE++\cite{liLearningEnhance2022} & TPAMI 2021 & 16.69 & 0.7903 & 0.1583 & 7.79 & 0.6081 & 0.4170 & 12.24 & 0.6992 & 0.2876 & 44.24 \\
    &PairLIE\cite{fuLearningSimple2023}  & CVPR 2023 & 14.87 & 0.7415 & 0.2686 & 8.53 & 0.6121 & 0.4084 & 11.70 & 0.6768 & 0.3385 & 80.13 \\
    &Retinexformer\cite{caiRetinexformerOnestage2023} & ICCV 2023 & \underline{19.47} & \underline{0.8331} & 0.1820 & 17.18 & 0.7846 & 0.2730 & \underline{18.32} & 0.8089 & 0.2275 & 40.53 \\
    &IAT\cite{Cui_2022_BMVC} & BMVC 2022 & 18.43 & 0.8025 & 0.1853 & 16.62 & 0.7671 & 0.2952 & 17.52 & 0.7848 & 0.2403 & 34.48 \\
    &LACT\cite{baekLuminanceawareColor2023b} & ICCV 2023 & 17.89 & 0.7635 & 0.2283 & 16.71 & 0.7600 & 0.3137 & 17.30 & 0.7617 & 0.2710 & 34.12 \\
    &SI-HDR\cite{leSingleImageHDR2023} & WACV 2023&  14.51 &0.7261&0.3265&16.33&0.7588&0.2480&15.42&0.7425&0.2872&36.67\\
    &MSLT++\cite{zhou4KResolutionPhoto2024} & WACV 2024 & 16.93 & 0.7898 & 0.1990 & 16.53 & 0.7676 & 0.3157 & 16.73 & 0.7787 & 0.2573 & 79.50 \\
    &RECNet\cite{liuRegionAwareExposure2024} & AAAI 2024 & 18.52 & 0.7954 & 0.1847 & 16.88 & 0.7773 & 0.2677 & 17.70 & 0.7864 & 0.2262 & 71.32 \\
    &CSEC\cite{liColorShift2024} & CVPR 2024 & 17.42 & 0.8042 & 0.1502 & 15.97 & 0.7656 & 0.2499 & 16.70 & 0.7849 & 0.2000 & 41.75 \\
    &CoTF\cite{liRealTimeExposure2024} & CVPR 2024 & 18.54 & 0.8299 & {0.1385} & 17.04 & \underline{0.7989} & 0.2359 & 17.79 & \underline{0.8144} & 0.1872 & 52.01 \\   
    &ExposureSlot\cite{jungExposureslotExposurecentric2025a} & CVPR 2025 & 18.99&0.8221&{0.1343}&16.55&0.7623&0.2595&17.77&0.7922&0.1969&41.23 \\
    \hline
    Event &NER-Net+\cite{liuNERNetSeeing2025} & TPAMI 2025 &  17.84&0.5480&0.4805& -&-&-&17.84&0.5480&0.4805&56.97 \\
    \hline
    \multirow{5}{*}{I+E} &UEVD~\cite{kimEventguidedDeblurring2022}& ECCV 2022&19.21&0.7965&0.2006&16.98&0.7471&0.3069&18.10&0.7718&0.2537&75.17 \\
    &EFNet~\cite{sunEventBasedFusion2022} & ECCV 2022&18.53&0.7760&0.1637&\underline{17.60}&0.7600&0.2517&18.07&0.7680&0.2077&43.44\\
    &REFID~\cite{sunUnifiedFramework2025} & TPAMI 2025 &11.82 &0.6026&0.6392&12.03&0.6141&0.5155&11.93&0.6084&0.5774&36.01\\
    &LIE\cite{jiangEventBasedLowIllumination2024} & TMM 2024 & 17.69&0.7948&\underline{0.1304}& 16.60&0.7020&0.2273& 17.15& 0.7484& 0.1789& 47.82\\
    &EvLight\cite{liangRobustEventguided2024} & CVPR 2024 & 18.55&0.7926&0.1372& {17.55} &0.7349& \underline{0.2047} & 18.05& 0.7637& \underline{0.1709} & \underline{33.21} \\
    \hline
    I+E &\textbf{Ours} & -& \textbf{20.06} & \textbf{0.8389} & \textbf{0.1249} & \textbf{18.06} & \textbf{0.8211} & \textbf{0.1736} & \textbf{19.03} & \textbf{0.8296} & \textbf{0.1501} & \textbf{32.25} \\
    \bottomrule
    \end{tabular}
    \begin{tablenotes}
        \item NER-Net+ is measured in grayscale and its results under overexposure are omitted due to poor performance and marked as ``-''.
        \item ``I+E'' denotes image-event fusion methods.
    \end{tablenotes}
    \end{threeparttable}
\end{table*}

\begin{figure*}[t]
    \centering
    \includegraphics[width=1.0\linewidth]{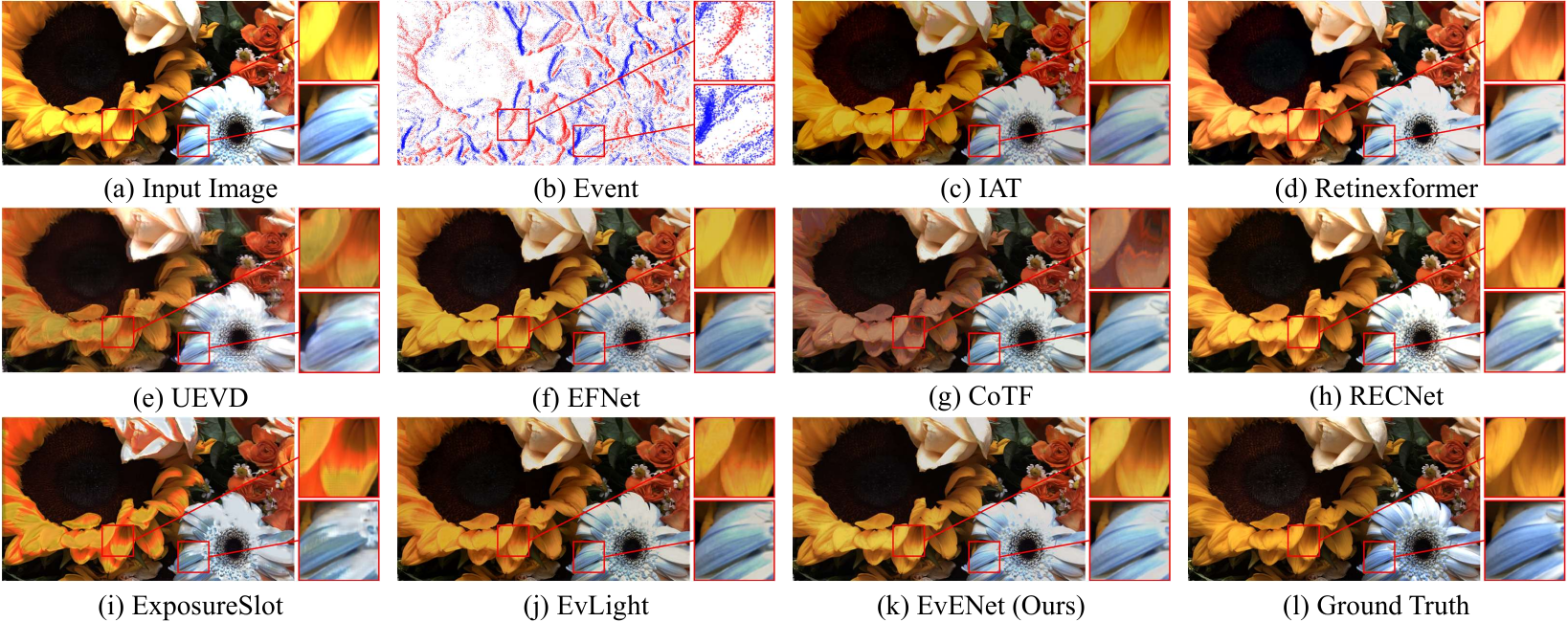}
    \caption{Qualitative comparison with state-of-the-art methods on the real-world REEC dataset. Our approach (k) allows for a better color fidelity and smooth chromatic transitions, avoiding unnatural color banding at illumination boundaries.}
    \label{fig:vis_REEC}
\end{figure*}

\subsection{Objective Loss Function}
Our objective function comprises four components: reconstruction loss, perceptual loss, Sobel loss, and the proposed similarity-based regularization loss:
\begin{equation}
    \label{eq:objective_loss}    
    \begin{aligned}    
    \mathcal{L} &= \mathcal{L}_{rec} + \lambda_p \mathcal{L}_{perceptual} + \lambda_s \mathcal{L}_{sobel} + \lambda_r \mathcal{L}_{sim}, \\
    \mathcal{L}_{sim}&=-\alpha\mathcal{L}_{{sim}_{modal}}+\beta\mathcal{L}_{{sim}_{scale}}
    \end{aligned}
\end{equation}
where hyperparameters $\lambda_p$, $\lambda_s$, $\lambda_r$, $\alpha$, and $\beta$ are set to 0.5, 0.05, 0.15, 0.3, and 0.1, respectively. 
We adopt Charbonnier loss $\mathcal{L}_{rec}$ to measure reconstruction errors and use pre-trained AlexNet features for $\mathcal{L}_{perceptual}$ to capture perceptual similarity. $\mathcal{L}_{sobel}$ preserves edge structures via Sobel gradient errors. 
The proposed similarity-based regularization loss $\mathcal{L}_{sim}$ comprises $\mathcal{L}_{{sim}_{modal}}$ and $\mathcal{L}_{{sim}_{scale}}$, which enforce cross-modal consistency and cross-scale feature decorrelation, respectively.

\section{Experiments}
\label{sec:experiment}
\textbf{Synthetic SICE-Event Dataset.}
To establish benchmark comparisons, we synthesize event data using the v2e emulator~\cite{Hu2021-v2e-cvpr-workshop-eventvision2021} on the SICE dataset, which contains the largest collection of real-world samples for exposure correction. The simulation adheres to v2e default parameters and generates 4,235 event-image pairs for training and 565 pairs for testing.

\textbf{Our REEC Dataset.} 
The training and testing scenarios of the REEC dataset are completely non-overlapping. We randomly partition the dataset into 89 training scenes and 22 testing scenes. Each scene contains 100 aligned event-image pairs captured under varying illumination conditions.

\textbf{DSEC Dataset.}
To evaluate the generalization of our method, we conduct cross-dataset experiments on the DSEC~\cite{Gehrig21ralDSEC} dataset. 
DSEC is a real-world driving dataset that records high-resolution event streams from an event camera at $640 \times 480$, along with synchronized images and LiDAR data.
We use the test set, which includes 12 scenarios, evaluating the first 30 frames in each according to the official settings.

\textbf{EventAid-D Dataset}
We evaluate generalization under real-world motion on EventAid-D~\cite{duanEventAidBenchmarking2025}, using seven of its sequences with available $1280\times720$ HDR references. These references are generated by fusing LDR images captured with cyclically alternating short, medium, and long exposures.

\textbf{Comparison Methods.}
Since existing event-guided methods are not specifically designed for exposure correction, we conduct a comprehensive comparison covering image-only, event-only, and image-event fusion methods.
For image-only baselines, we include single image HDR reconstruction SI-HDR~\cite{leSingleImageHDR2023}, low-light enhancement methods (ZeroDCE++~\cite{liLearningEnhance2022}, PairLIE~\cite{fuLearningSimple2023}, Retinexformer~\cite{caiRetinexformerOnestage2023}), and exposure correction methods (IAT~\cite{Cui_2022_BMVC}, LACT~\cite{baekLuminanceawareColor2023b}, MSLT++~\cite{zhou4KResolutionPhoto2024}, RECNet~\cite{liuRegionAwareExposure2024}, CSEC~\cite{liColorShift2024}, CoTF~\cite{liRealTimeExposure2024}, and ExposureSlot~\cite{jungExposureslotExposurecentric2025a}).
For event-only methods, we adopt NER-Net~\cite{liuNERNetSeeing2025}.
For image-event fusion methods, we consider representative approaches from  deblurring (UEVD~\cite{kimEventguidedDeblurring2022}, EFNet~\cite{sunEventBasedFusion2022}, and REFID~\cite{sunUnifiedFramework2025}) and low-light enhancement (LIE~\cite{jiangEventBasedLowIllumination2024}, EvLight~\cite{liangRobustEventguided2024}).
All methods are retrained on our REEC dataset for fair comparison.

\textbf{Implementation Details.} 
Our experiments are conducted in PyTorch on a single RTX 3090 GPU. To ensure fair comparison, we retrain state-of-the-art methods on the SICE-Event and REEC datasets using their official implementations. We adopt Adam optimizer with a learning rate of 1e-4, training for 100k iterations with batch size 8 and patch size 256$\times$256.

\begin{figure*}[!t]
    \centering
    \includegraphics[width=1.0\linewidth]{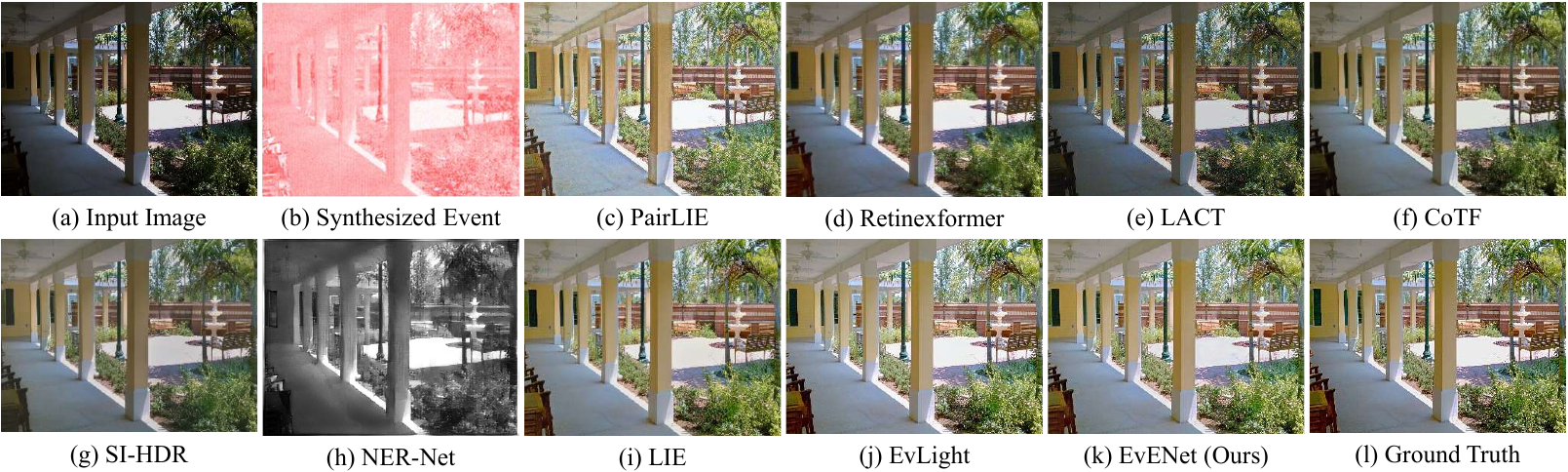}
    \caption{Qualitative comparison with state-of-the-art methods on the synthetic SICE-Event dataset. Our approach (n) EvENet shows a better ability to preserve the details in both under- and over-exposed regions.}
    \label{fig:vis_sice}
\end{figure*}

\begin{table}[!t]
    \centering
    \begin{threeparttable}
        \caption{Quantitative comparison with state-of-the-art methods on the synthetic SICE-Event dataset}
        \label{tab:result_sice}
        \begin{tabular}{c|c|cccc}
            \toprule
            \multirow{2}{*}{Method} &\multirow{2}{*}{Input}& \multicolumn{4}{c}{SICE-Event}\\
            &&PSNR$\uparrow$ & SSIM$\uparrow$ &LPIPS$\downarrow$  & ILNIQE$\downarrow$ \\
            \hline
            NER-Net+\cite{liuNERNetSeeing2025}&Event&11.05&0.3952&0.5439&33.7110\\ \cline{1-6}
            \hline
            SI-HDR\cite{leSingleImageHDR2023} &\multirow{11}{*}{Image}& 19.47 & 0.7602 & 0.2290 & 35.8153\\ 
            ZeroDCE++\cite{liLearningEnhance2022}&  & 13.21 & 0.6740 & 0.2877 & 31.6456 \\
            PairLIE\cite{fuLearningSimple2023} &&14.22 & 0.7100 & 0.2540 & 31.3337 \\
            Retinexformer\cite{caiRetinexformerOnestage2023} && \underline{21.10} & \underline{0.8297} & 0.1772 & 31.5629 \\            
            IAT\cite{Cui_2022_BMVC}&& 17.00 & 0.6893 & 0.2532 & 31.5520 \\
            LACT\cite{baekLuminanceawareColor2023b}  && 20.35 & 0.7733 & 0.1870 & 31.9439 \\
            MSLT++\cite{zhou4KResolutionPhoto2024}&& 19.98& 0.7467 & 0.2450 & 34.1134 \\
            RECNet\cite{liuRegionAwareExposure2024}&& 18.67 & 0.6422 & 0.2185 & 29.1434 \\
            CSEC\cite{liColorShift2024} && 19.27 & 0.7726 & 0.1799 & 29.8040 \\
            CoTF\cite{liRealTimeExposure2024} && 20.38 & 0.8009 & 0.1913 & 29.5492 \\
            ExposureSlot\cite{jungExposureslotExposurecentric2025a} &&19.23& 0.6506 & 0.1734& 27.1763\\ \cline{1-6}
            \hline
            UEVD~\cite{kimEventguidedDeblurring2022} &\multirow{5}{*}{I+E} &16.65&0.4257&0.2215&28.2846\\ 
            EFNet~\cite{sunEventBasedFusion2022} &&20.13&0.7971&0.1626&27.7284 \\
            REFID~\cite{sunUnifiedFramework2025} &&15.26&0.4977&0.6925&67.3882 \\
            LIE\cite{jiangEventBasedLowIllumination2024} && 20.96 & 0.8275 & 0.1561& 37.3633\\
            EvLight\cite{liangRobustEventguided2024} &&20.28 & 0.7548 & \underline{0.1346} & \underline{26.8193}\\ \cline{1-6}
            \hline
            \textbf{Ours} & I+E&  \textbf{22.07} & \textbf{0.8344} & \textbf{0.1168} & \textbf{26.2190} \\ 
            \bottomrule
        \end{tabular}
    \end{threeparttable}
\end{table}

\textbf{Evaluation Metrics.}
We employ five metrics for evaluation: PSNR, SSIM, LPIPS, ILNIQE, and PIQE to comprehensively assess the performance of each method.

\subsection{Quantitative Experiments}
\label{subsec:quantitative_exp}
We conduct quantitative experiments across diverse datasets to comprehensively evaluate the performance of our method.
Tab.~\ref{tab:result_REEC} presents results on REEC, with metrics reported separately for under- and over-exposed samples.
Our method consistently outperforms prior approaches across all evaluation metrics.
We further analyze the performance across different categories of methods. Image-only methods generally achieve competitive results under under-exposed conditions, as they can leverage visible structures for enhancement. For instance, Retinexformer achieves strong performance with 19.47 PSNR under under-exposure. However, their performance degrades noticeably under over-exposure due to the lack of recoverable information in saturated regions.
For the event-only methods, when the event stream is sparse or fails to provide sufficient contrast-triggered responses, the reconstruction quality degrades significantly and does not encode sufficient constraints for recovering absolute intensity or color information. 
As a result, such methods fail to reconstruct high-quality images when event signals are weak or uninformative.
Image-event fusion methods achieve more balanced performance by combining complementary modalities. Video reconstruction methods such as UEVD demonstrate improved robustness compared to single-modality approaches. However, their performance is still limited by suboptimal fusion strategies.
In contrast, our method achieves the best performance across both under- and over-exposed subsets, demonstrating its ability to effectively integrate event-based temporal cues with image-based intensity and color information. Notably, our method shows a clear advantage in over-exposed regions, indicating its superior capability in recovering details from saturated areas.

Tab.~\ref{tab:result_sice} reports results on the synthetic SICE-Event dataset.
From the results, there exist consistent trends as Tab.~\ref{tab:result_REEC}. Event-only methods again perform poorly, confirming that relying solely on event streams is insufficient for exposure correction. Image-only methods show competitive performance but remain limited in handling extreme illumination variations.
Among image-event fusion methods, LIE and EvLight achieve relatively strong performance. This suggests that effective cross-modal interaction is critical for optimal performance.
In contrast, EvENet achieves superior results by jointly modeling event-image representations.

\begin{table}[!t]
    \centering
    \begin{threeparttable}
        \caption{Quantitative comparison with state-of-the-art methods on the real-world DSEC and EventAid-D datasets.}
        \label{tab:result_dsec_eventaid}
        \begin{tabular}{@{}c@{}|@{ }c@{ }|@{ }c@{ }c@{ }|c@{ }c@{ }c}
            \toprule
            \multirow{2}{*}{Method} &\multirow{2}{*}{Task} & \multicolumn{2}{c|}{DSEC}& \multicolumn{3}{c}{EventAid-D} \\
            && PIQE$\downarrow$ & ILNIQE$\downarrow$ &PSNR$\uparrow$ & SSIM$\uparrow$ &LPIPS$\downarrow$\\
            \hline
            ZeroDCE++\cite{liLearningEnhance2022}& \multirow{6}{*}{LLE} &27.84&20.62 & 18.14 & 0.6332 & 0.3356 \\
            PairLIE\cite{fuLearningSimple2023} & &65.97&39.29& 10.39 & 0.4274 & 0.4676 \\
            Retinexformer\cite{caiRetinexformerOnestage2023} & &26.77&26.80 & 16.62 & 0.5781 & 0.2913 \\
            NER-Net+\cite{liuNERNetSeeing2025}&&40.90&56.93 & 12.77 & 0.3795 & 0.4794 \\
            LIE\cite{jiangEventBasedLowIllumination2024} &&21.00&\underline{19.45} & 13.19 & 0.4635 & 0.4206 \\
            EvLight\cite{liangRobustEventguided2024} &&17.10&23.69& \underline{20.94} & 0.6788 & \underline{0.2238} \\
            \hline
            SI-HDR\cite{leSingleImageHDR2023} &HDR&29.29&32.47& 11.62 & 0.4696 & 0.4402 \\
            \hline
            UEVD~\cite{kimEventguidedDeblurring2022} &\multirow{3}{*}{Others} &64.13&32.79& 19.37 & 0.6781 & 0.2677 \\
            EFNet~\cite{sunEventBasedFusion2022} &&18.89&27.35& 18.46 & 0.6285 & 0.2468 \\
            REFID~\cite{sunUnifiedFramework2025} &&26.18&59.67& 10.95 & 0.3794 & 0.5697 \\
            \hline
            IAT\cite{Cui_2022_BMVC}& \multirow{7}{*}{EC}&21.43&27.26& 17.37 & 0.6325 & 0.3577\\
            LACT\cite{baekLuminanceawareColor2023b} &&22.68&23.23& 19.22 & \textbf{0.7329} & 0.2760 \\
            MSLT++\cite{zhou4KResolutionPhoto2024}&&58.09&37.44& 14.44 & 0.5553 & 0.3662 \\
            RECNet\cite{liuRegionAwareExposure2024}&&55.28&29.94& 17.78 & 0.6440 & 0.3041 \\
            CSEC\cite{liColorShift2024} &&23.50&19.76& 12.84 & 0.5392 & 0.4127 \\
            CoTF\cite{liRealTimeExposure2024} &&28.60&20.87& 13.44 & 0.5253 & 0.3347\\
            ExposureSlot\cite{jungExposureslotExposurecentric2025a} &&21.63&19.57& 14.02 & 0.4991 & 0.3555 \\
            \hline
            \textbf{Ours} & EC&\textbf{16.87}&\textbf{19.42}& \textbf{21.30} & \underline{0.6847} & \textbf{0.2211} \\ 
            \bottomrule
        \end{tabular}
        \begin{tablenotes}
            \item LLE: low-light enhancement; HDR: HDR reconstruction; EC: exposure correction; Others: event-based reconstruction and deblurring.
        \end{tablenotes}
    \end{threeparttable}
    \vspace{-2em}
\end{table}

\begin{figure*}[!t]
    \centering\
    \includegraphics[width=.9\linewidth]{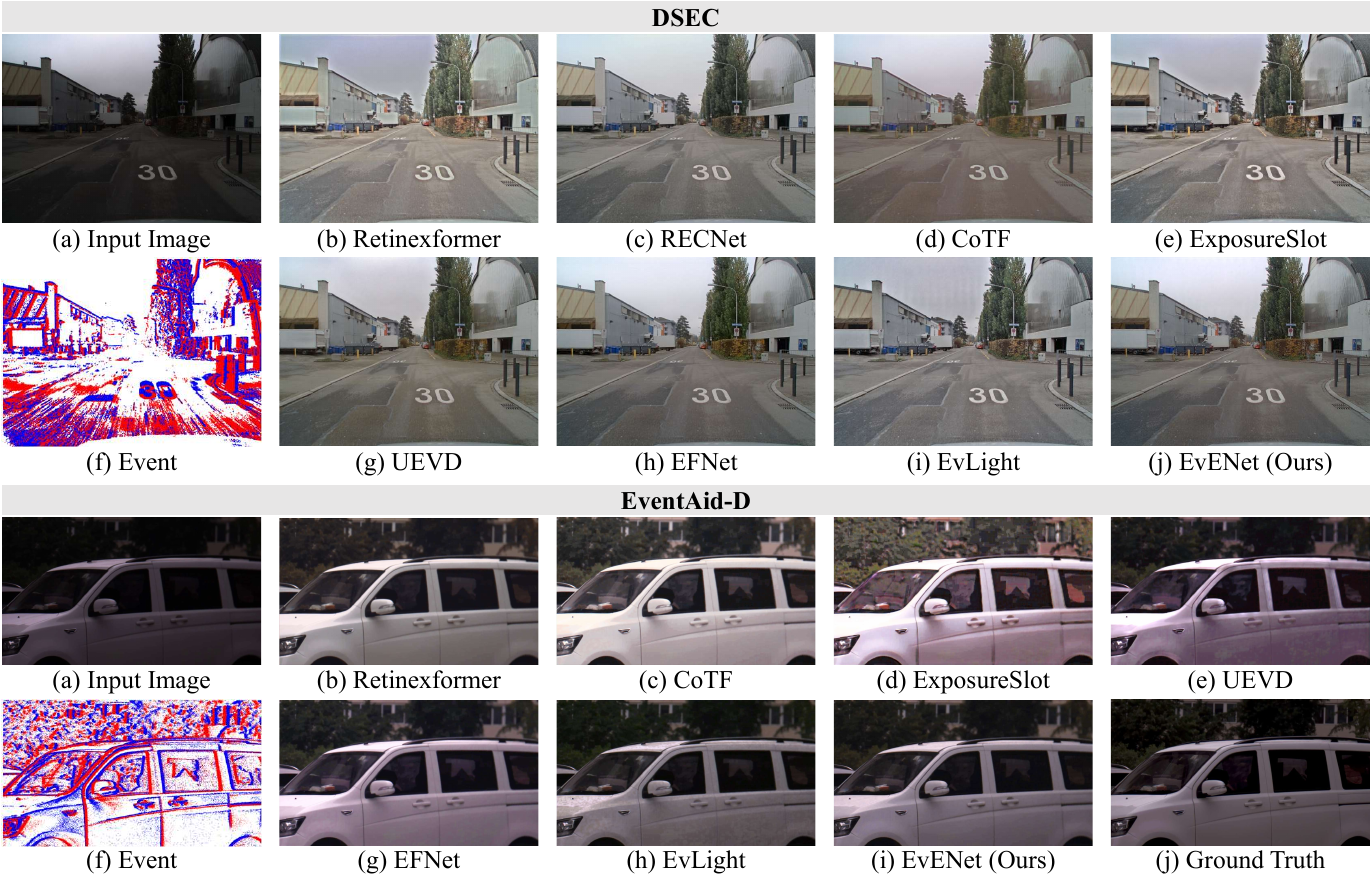}
    \caption{Qualitative comparison on the real DSEC and EventAid-D datasets. Our EvENet improves detail visibility while preserving colors and scene contrast.}
    \label{fig:vis_dsec_eventaid}
    \vspace{-1em}
\end{figure*}

Tab.~\ref{tab:result_dsec_eventaid} presents cross-dataset evaluations on DSEC and EventAid-D. The favorable no-reference results on DSEC indicate that EvENet maintains image quality in real driving scenes, while the reference-based results on EventAid-D show that this performance extends to reconstruction fidelity and perceptual similarity under real-world motion. These evaluations support the generalization of our exposure-aware fusion framework across different motion conditions.

Additionally, the inference efficiency at 1280$\times$720 resolution is reported in Tab.~\ref{tab:costs}. Compared with representative high-performing methods, including image-based Retinexformer~\cite{caiRetinexformerOnestage2023} and event-based EvLight~\cite{liangRobustEventguided2024} and UEVD~\cite{kimEventguidedDeblurring2022}, our EvENet achieves higher PSNR while being faster and more memory-efficient.

\begin{table}[t]
    \centering
    \caption{Comparisons of computational costs with event-based methods on the REEC dataset at 1280$\times$720 resolution.}
    \label{tab:costs}
    \begin{tabular}{l
        S[table-format=3.2,mode=text]
        S[table-format=3.2,mode=text]
        S[table-format=5.0,mode=text]
        S[table-format=2.2,mode=text]}
    \toprule
    Method & \multicolumn{1}{c}{\makecell{Params\\(M)}} & \multicolumn{1}{c}{\makecell{Inference Time\\(ms)}} & \multicolumn{1}{c}{\makecell{Memory \\ (MB)}} & \multicolumn{1}{c}{PSNR} \\
    \hline
    ZeroDCE++~\cite{liLearningEnhance2022}  & 0.01 & 47.35 & 425 & 12.24 \\    
    PairLIE~\cite{fuLearningSimple2023} & 0.34 & 83.27 & 1261 & 11.70\\
    Retinexformer~\cite{caiRetinexformerOnestage2023} & 1.61 & 323.98 & 3941&18.32\\
    SI-HDR~\cite{leSingleImageHDR2023}  & 82.22 & 125.84 & 3290& 15.42 \\
    IAT~\cite{Cui_2022_BMVC} & 0.09 & 75.47 & 1329 &17.52\\
    LACT~\cite{baekLuminanceawareColor2023b} & 2.25 & 378.97 & 22734&17.30\\
    MSLT++~\cite{zhou4KResolutionPhoto2024} & 0.01 & 95.25 & 428&16.73\\
    RECNet~\cite{liuRegionAwareExposure2024} & 0.04 & 272.84 & 1267 &17.70\\
    CSEC~\cite{liColorShift2024} & 1.36 & 836.39 & 6237 & 16.70\\
    CoTF~\cite{liRealTimeExposure2024} & 0.31 & 174.98 & 1975 & 17.79\\
    ExposureSlot~\cite{jungExposureslotExposurecentric2025a} & 2.01 & 129.16 & 2097 & 17.77\\
    \midrule
    {UEVD~\cite{kimEventguidedDeblurring2022}}&27.88&737.15&13810&18.10\\
    {EFNet~\cite{sunEventBasedFusion2022}} &8.48&605.30&5438&18.07\\
    {REFID~\cite{sunUnifiedFramework2025}}&15.93&458.36&3468&11.93\\
    {NER-Net+\cite{liuNERNetSeeing2025}} & 19.36 & 261.30 & 5763& 17.84\\
    LIE~\cite{jiangEventBasedLowIllumination2024} & 220.03 & 2767.90 & 12615 & 17.15 \\
    EvLight~\cite{liangRobustEventguided2024} & 22.73 & 594.74 & 5059 & 18.05 \\
    \midrule
    EvENet (Ours) & 10.69 & 280.83& 3388 & 19.03\\
    \bottomrule
    \end{tabular}
    \vspace{-1em}
\end{table}

\subsection{Qualitative Experiments}
We conduct qualitative experiments on the REEC and SICE-Event datasets, as well as cross-dataset evaluations on the real-world DSEC dataset, to thoroughly assess the generalization capability of our proposed method.

\noindent\textbf{Visual Evaluation on our real-world REEC.}
Fig.~\ref{fig:vis_REEC} presents the visual comparisons on the REEC dataset, which contains real-world scenes with non-uniform illumination, including both under- and over-exposed regions.
Representative exposure correction methods, such as ExposureSlot~\cite{jungExposureslotExposurecentric2025a} and CoTF~\cite{liRealTimeExposure2024}, exhibit noticeable color shifts at illumination boundaries.
Retinexformer~\cite{caiRetinexformerOnestage2023} reconstructs images with blurred details, while the event-based low-light enhancement method EvLight~\cite{liangRobustEventguided2024} introduces visible noise. 
The event-guided restoration methods, UEVD~\cite{kimEventguidedDeblurring2022}, suffer from noticeable motion blur, whereas the deblurring approach EFNet~\cite{sunEventBasedFusion2022} shows lower color fidelity.
In contrast, our method, EvENet, preserves clear edges and natural color fidelity.

\noindent\textbf{Visual Evaluation on synthetic SICE-Event.}
Fig.~\ref{fig:vis_sice} presents visual comparisons on the synthetic SICE-Event dataset. 
Most competing methods suffer from unnatural color shifts and noticeable artifacts, whereas our method consistently delivers superior detail preservation and vivid color reconstruction. 

\noindent\textbf{Visual Evaluation on real datasets.}
Fig.~\ref{fig:vis_dsec_eventaid} illustrates the visual comparisons on DSEC and EventAid-D, showing the importance of balancing visibility enhancement with appearance preservation.
In the DSEC, EvENet improves the visibility of the road and buildings while retaining contrast between shaded and illuminated regions. In the EventAid-D, several competing methods introduce color shifts or excessive brightening, whereas EvENet produces vehicle and background appearances closer to the reference.

\subsection{Ablation Studies}
\subsubsection{Ablations on Design Choices}
\begin{figure}[!t]
    \centering
    \includegraphics[width=1.0\linewidth]{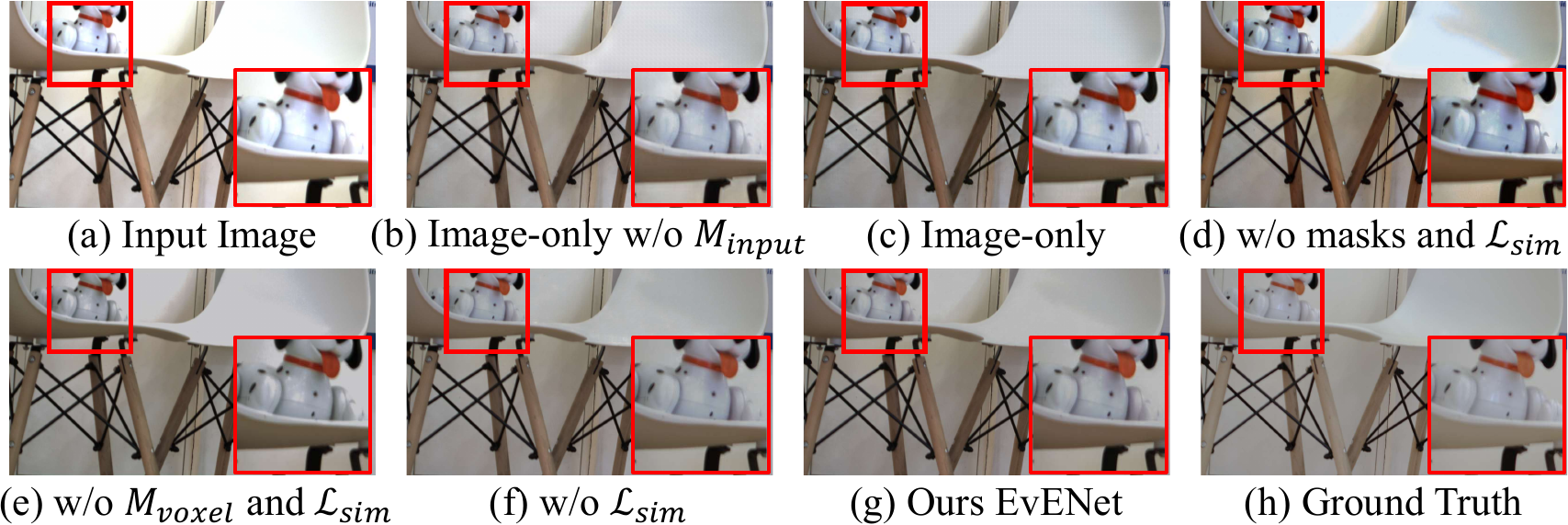}
    \caption{Visual comparison of ablation results on the REEC dataset.}
    \label{fig:ablations_reec_vis}   
    \vspace{-1em}
\end{figure}

To systematically examine the effect of different design choices, we conduct ablation studies on the REEC dataset.
Tab.~\ref{tab:ablations_component} reports the performance under various configurations, with corresponding visual results shown in Fig.~\ref{fig:ablations_reec_vis}. 
The image-only comparisons show that $\mathbf{M}_{input}$ improves reflection suppression by selectively weighting image features. Incorporating event inputs without mask guidance yields inferior results, highlighting the importance of selective fusion in exploiting complementary event information. Progressively enabling $\mathbf{M}_{input}$ and $\mathbf{M}_{voxel}$ improves reconstruction quality by emphasizing informative image features and regulating the contribution of event features, respectively.
Furthermore, removing the dual-objective similarity loss degrades performance, as this loss facilitates the preservation of high-frequency details captured by event streams while suppressing reflection artifacts.

\subsubsection{Ablations on Fusion Objectives}
To investigate the impact of different fusion objectives, we conduct an ablation study, as shown in Tab.~\ref{tab:ablations_mi_estimator}.
MINE~\cite{belghaziMutualInformation2018} exhibits training instability and fails to converge. It primarily suffers from high gradient variance, particularly when dealing with heterogeneous modalities such as images and events.
InfoNCE~\cite{oordRepresentationLearning2019} improves performance when only a single constraint is applied, but its reliance on explicit negative sampling introduces additional optimization complexity and memory overhead.
In contrast, our cosine similarity-based formulation achieves the best overall performance among the evaluated fusion objectives when both cross-modal alignment and inter-scale similarity regularization are jointly applied. This indicates that explicit density estimation or contrastive sampling is not necessary for effective fusion in event-based exposure correction, and that a simple similarity-based constraint is sufficient to guide representation learning.

\begin{table}[t]
    \centering
    \caption{Quantitative ablations of EvENet on the REEC
    dataset.}
    \label{tab:ablations_component}
    \setlength{\tabcolsep}{4pt}
    \renewcommand{\arraystretch}{1.0}
    \begin{threeparttable}
    \begin{tabular}{lrr}
        \toprule
        Configuration & PSNR$\uparrow$ & SSIM$\uparrow$ \\
        \midrule
        \textbf{Full EvENet} & \textbf{19.03} & \textbf{0.8296} \\
        \midrule
        w/o $\mathbf{M}_{input}$, $\mathbf{M}_{voxel}$, $\mathcal{L}_{sim}$ & 16.70 & 0.7091 \\
        w/o $\mathbf{M}_{voxel}$, $\mathcal{L}_{sim}$ & 17.79 & 0.7854 \\
        w/o $\mathcal{L}_{sim}$ & 18.89 & 0.8027 \\
        \midrule
        Image-only\tnote{*} & 18.37 & 0.7498 \\
        Image-only\tnote{*} w/o $\mathbf{M}_{input}$ & 18.12 & 0.7365 \\
        \bottomrule
    \end{tabular}
    \begin{tablenotes}[flushleft]
        \footnotesize
        \item[*] Image-only removes event inputs, $\mathbf{M}_{voxel}$, and $\mathcal{L}_{sim}$. Ablated masks are set to unity.
    \end{tablenotes}
    \end{threeparttable}
    \vspace{-1em}
\end{table}

\begin{table}[!t]
    \centering
    \begin{threeparttable}
    \caption{Quantitative ablations on similarity-based fusion objectives.}
    \label{tab:ablations_mi_estimator}
    \begin{tabular}{c@{}|@{ }c|@{}c@{}c@{}|c@{}r}
    \toprule
    Fusion & Memory & 
    \multirow{2}{*}{$\mathcal{L}_{{sim}_{scale}}$} & \multirow{2}{*}{$\mathcal{L}_{{sim}_{modal}}$} & \multirow{2}{*}{PSNR} & \multirow{2}{*}{SSIM} \\
    Objective & Overhead (MB)& & & & \\
    \midrule
    MINE\cite{belghaziMutualInformation2018} & +5510 & \ding{51} & \ding{51} & 
    \multicolumn{2}{c}{\textit{Training diverged}} \\
    \midrule
    \multirow{3}{*}{InfoNCE\cite{oordRepresentationLearning2019}} & \multirow{3}{*}{+2670} &
    \ding{51} & \ding{55} & 18.18 & 0.7612\\
    & & \ding{55} & \ding{51} & 17.73 & 0.7515\\
    & & \ding{51} & \ding{51} & 17.56 & 0.7454\\
    \midrule
    \multirow{3}{*}{Ours} & \multirow{3}{*}{+0} & \ding{51} & \ding{55} & 18.11 & 0.7791\\
    & & \ding{55} & \ding{51} & 18.36 & 0.7820\\
    & & \ding{51} & \ding{51} & 19.03 & 0.8296\\
    \bottomrule
    \end{tabular}
    \begin{tablenotes}
    \footnotesize
    \item * Memory Overhead denotes additional GPU memory usage compared with the DotProduct baseline.
    \end{tablenotes}
    \end{threeparttable}
\end{table}

\subsubsection{Ablations on the Window Lengths}
To investigate the effect of temporal window length in event voxelization, we conduct an ablation study by varying the window size. The temporal window $W(m)$ is defined in Eq.~\ref{eq:window_define}, where $T_c$ denotes the center of the RGB exposure, and $\Delta T = m \cdot T_E$ is parameterized as a multiple of the exposure duration $T_E$.
\begin{equation}
    \label{eq:window_define}
    W(m) = [T_c - \Delta T, T_c + \Delta T]
\end{equation}
We evaluate different window scales with $m \in \{0.25, 0.5, 0.75, 1.00, 1.50, 2.00\}$. Note that $m=0.5$, corresponding to a window length equal to the RGB exposure time, is adopted as the default setting in other experiments. Quantitative results are summarized in Tab.~\ref{tab:ablations_windows}.
From the results, we observe that a moderate temporal window ($m=1.00$, i.e., $2T_E$) achieves the best performance in terms of PSNR and SSIM. This indicates that incorporating events beyond the exact exposure interval can provide complementary temporal cues beneficial for exposure correction.
When the window is too short (e.g., $m=0.25$), the event voxel becomes sparse, leading to insufficient temporal information and degraded reconstruction quality. In contrast, excessively large windows (e.g., $m \geq 1.50$) aggregate events over extended motion trajectories, which introduces temporal inconsistency and noise, thereby harming performance.

Notably, a window covering approximately half of the neighboring frames' exposure (i.e., extending beyond the current frame by $\frac{1}{2}T_E$ on each side) yields consistent improvements. This suggests that limited cross-frame temporal aggregation helps capture additional brightness variations without introducing significant motion-induced artifacts.
Overall, these results validate the trade-off that an appropriate temporal window balances event density and temporal consistency, which is critical for effective event-guided exposure correction.

\begin{table}[t]
    \centering    
    \caption{Quantitative ablations of diverse window lengths in event representation.}
    \label{tab:ablations_windows}
    \begin{tabular}{cc|ccc}
    \toprule
    $m$ & Window Length & PSNR & SSIM & LPIPS \\
    \midrule
    0.25 &$\text{0.5} \cdot T_E$&18.42 &0.8106 &0.1567 \\
    0.50 &$\text{1.0} \cdot T_E$&19.03 &0.8296 &0.1501\\
    0.75 &$\text{1.5} \cdot T_E$&18.28 &0.8171 &0.1584 \\
    1.00 &$\text{2.0} \cdot T_E$&19.12 &0.8334 &0.1523 \\
    1.50 &$\text{3.0} \cdot T_E$&18.56 &0.8218 &0.1565 \\
    2.00 &$\text{4.0} \cdot T_E$&18.53 &0.8167 &0.1562 \\
    \bottomrule
    \end{tabular}
    \vspace{-1em}
\end{table}

\subsubsection{Ablations on the Number of Bins}
We investigate the effect of the number of bins ($\#\text{bins}$) used in constructing event voxel representations. Specifically, we evaluate different bin configurations under a fixed temporal window $\Delta T$, which is set to match the exposure duration of the corresponding RGB image, where the event stream is uniformly partitioned along the temporal dimension.
A smaller $\#\text{bins}$ results in coarser temporal discretization, where each voxel aggregates a larger number of events within a wider temporal interval. Conversely, a larger $\#\text{bins}$ provides finer temporal sampling, producing a more temporally resolved but sparser event representation per bin.
We evaluate commonly used settings, including $\#\text{bins} \in \{5, 6, 10, 16, 32, 36\}$, and report results in Tab.~\ref{tab:ablations_bins}.
From a computational perspective, the number of parameters remains nearly unchanged across different bin settings, while inference time shows a slight variation due to differences in tensor dimensionality and memory access patterns.

\begin{figure}[t]
    \centering
    \includegraphics[width=1.0\linewidth]{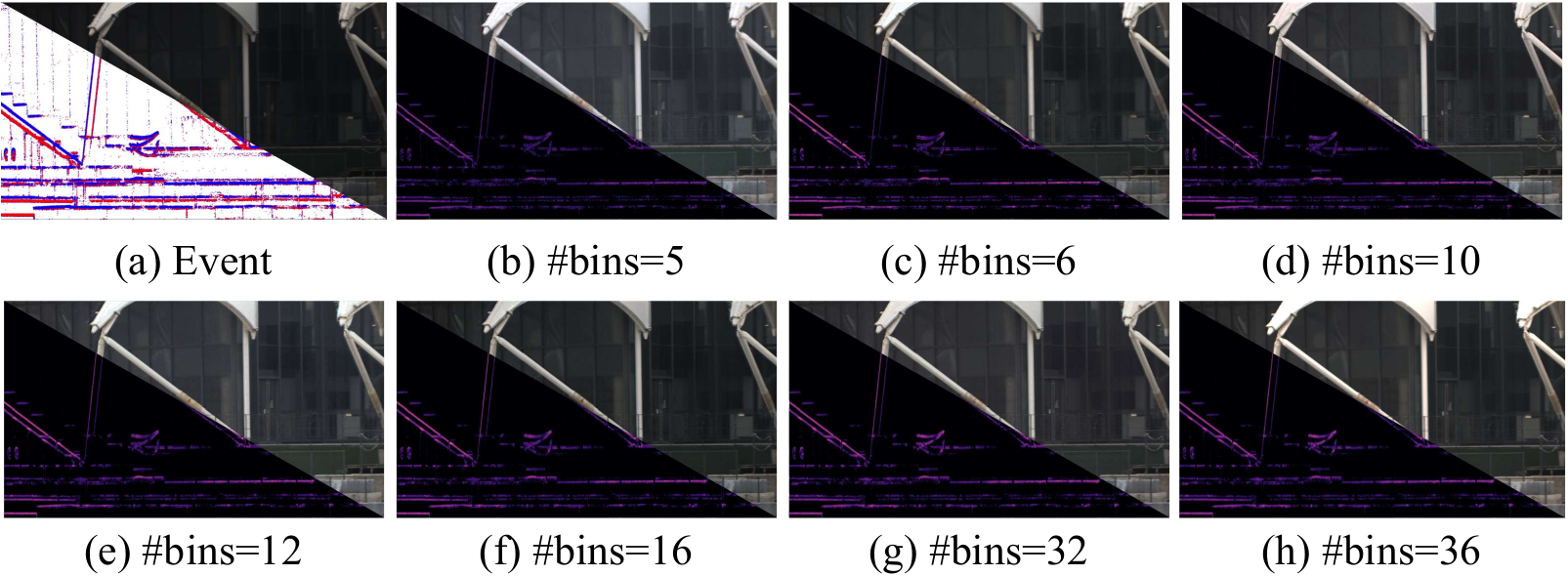}
    \caption{Visualization of predicted image and event voxel representations with varying numbers of temporal bins.}
    \label{fig:vis_bins_image_voxel}
\end{figure}

We further visualize voxel representations under different bin configurations in Fig.~\ref{fig:vis_bins_image_voxel}. As $\#\text{bins}$ increases, the event voxel becomes more temporally localized, preserving finer motion-related structures within each bin, while each channel contains fewer aggregated events.
Quantitatively, we observe that the smallest bin setting achieves the best PSNR and SSIM. It suggests that, under a fixed exposure interval aligned with RGB frames, coarser temporal aggregation may provide more robust structural cues for reconstruction. In contrast, finer binning does not consistently translate into performance gains, potentially due to increased sparsity per channel.
Our default setting $\#\text{bins}=32$ achieves a favorable trade-off, yielding competitive PSNR and LPIPS, while maintaining stable perceptual quality across scenarios.
Overall, there exists a clear trade-off between temporal resolution and reconstruction performance, as well as computational efficiency.

\begin{table}[t]
    \centering    
    \caption{Quantitative ablations of number of bins in event voxel representation.}
    \label{tab:ablations_bins}
    \begin{tabular}{c|ccccc}
    \toprule    
    \multirow{2}{*}{\text{\#Bins}} & \text{\#Params} & Inference Time & \multirow{2}{*}{PSNR} & \multirow{2}{*}{SSIM} &\multirow{2}{*}{LPIPS} \\
     & (M) & (ms)&& \\
    \hline
    5  &10.688&301.06&\textbf{19.13}&\textbf{0.8391}&0.1532\\
    6  &10.688&294.42&18.74&\underline{0.8368}&0.1575\\
    10 &10.689&288.17&18.34&0.8029&0.1680\\
    16 &10.691&277.11&18.99&0.7719&\textbf{0.1494}\\
    32 &10.695&280.83&\underline{19.03}&0.8296&\underline{0.1501}\\
    36 &10.697&277.59&18.54&0.8329&0.1620\\
    \bottomrule
    \end{tabular}
\end{table}

\subsubsection{PCA Analysis}
\begin{figure}[!t]
    \centering
    \includegraphics[width=1.0\linewidth]{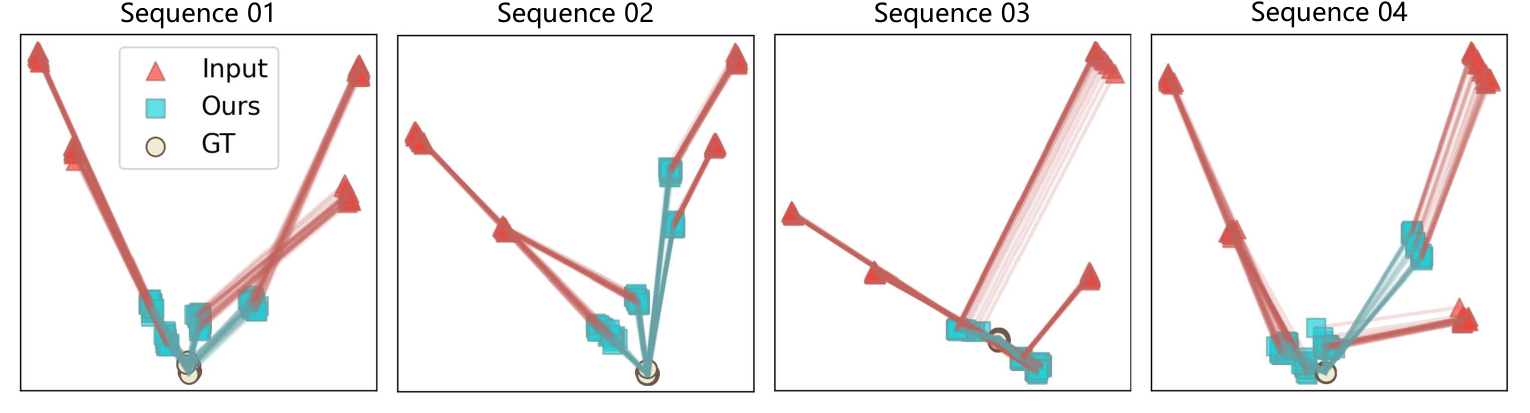}
    \caption{PCA visualizations for parts of REEC testing set.}
    \label{fig:ablations_PCA}
    \vspace{-1em}
\end{figure}

In addition, we visualize the statistical distribution of features using PCA, as shown in Fig.~\ref{fig:ablations_PCA}. After processing by our method, the ill-exposed inputs cluster closely with their corresponding ground truth samples, demonstrating that our approach effectively aligns exposure representations across different scenes.

\begin{figure}[!t]
    \centering
    \includegraphics[width=1.0\linewidth]{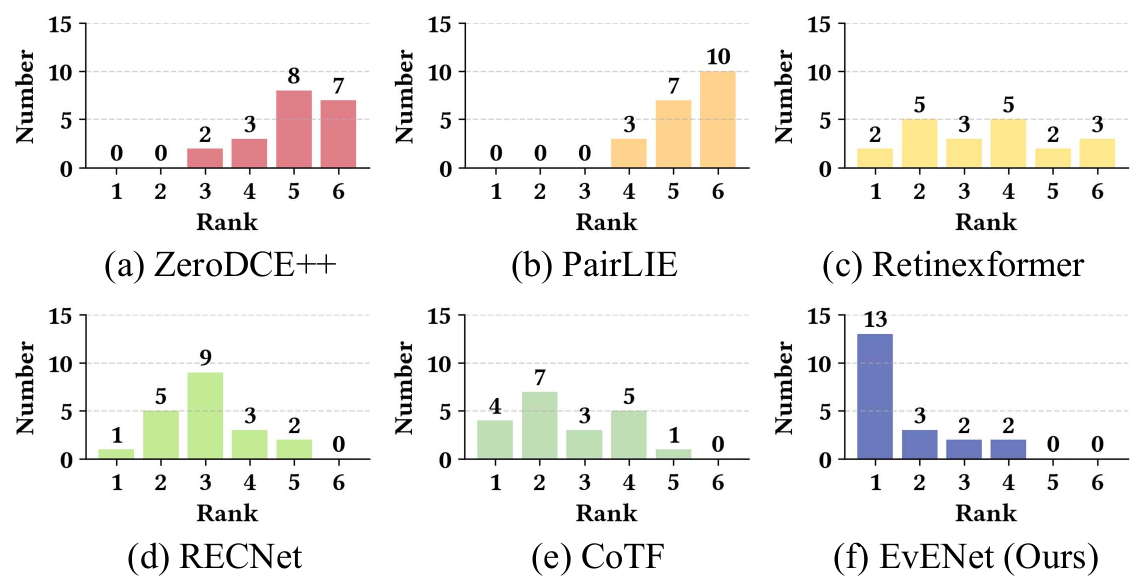}
    \caption{Results of the user study on the REEC dataset. Rank 1 is the best, and 6 is the worst. Our method achieves the highest first-place rankings.}
    \label{fig:user_study_rank}
    \vspace{-1em}
\end{figure}

\subsection{User Study} 
To comprehensively validate the performance, we conduct a user study comparing our method with five state-of-the-art methods from Tab.~\ref{tab:result_REEC} on the REEC dataset. Ten volunteers rank 20 randomly selected test images from 1 (best) to 6 (worst) under identical conditions (room, display, lighting). As shown in Fig.~\ref{fig:user_study_rank}, our method achieves 13 first-place and 3 second-place rankings, demonstrating clear superiority over competing methods.

\begin{figure}[!t]
    \centering
    \includegraphics[width=1.0\linewidth]{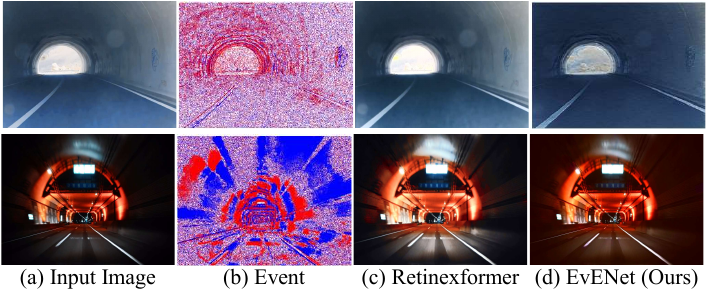}
    \caption{Qualitative comparison on the real-world challenging scenarios.}
    \label{fig:vis_challenging_R2}
    \vspace{-1em}
\end{figure}

\subsection{Limitations and Analysis}
To further evaluate performance under high-speed conditions, we collect several challenging sequences from online sources.
The visual comparison is shown in Fig.~\ref{fig:vis_challenging_R2}.
In such extreme illumination variation and ultra-high-speed scenarios, existing image-only method exhibits noticeable artifacts.
In top row of Fig.~\ref{fig:vis_challenging_R2}, our approach (d) effectively balances the restoration of underexposed regions inside the tunnel and the recovery of overexposed regions outside the tunnel exit.
However, in bottom row of Fig.~\ref{fig:vis_challenging_R2}, it still struggles to recover severe motion blur when relying on synthetic event inputs.

\section{Conclusion}
\label{sec:conclusion}
We introduce REEC, a comprehensive high-resolution real-world dataset for event-based exposure correction, constructed using a precisely calibrated hybrid-camera system. To leverage REEC, we propose EvENet, an event-guided network that employs exposure-aware cross-modal masks and a similarity-driven fusion strategy to effectively integrate RGB images and event streams. Extensive experiments on both synthetic and real-world datasets demonstrate the effectiveness of REEC and validate the superior performance of EvENet compared to state-of-the-art exposure correction methods.

\bibliographystyle{IEEEtran}
\bibliography{main}

\end{document}